%% file: ICCV19.tex
\documentclass[10pt,twocolumn,letterpaper]{article}

\usepackage{iccv}
\usepackage{times}
\usepackage{epsfig}
\usepackage{graphicx}
\usepackage{amsmath}
\usepackage{amssymb}
\usepackage{mathtools}
\usepackage{xcolor}
\usepackage[caption=false]{subfig}
\usepackage{multirow}

\usepackage{authblk}

\makeatletter
\renewcommand\AB@affilsepx{\hspace{1em}\protect\Affilfont}
\makeatother

\graphicspath{ {images/} }

\usepackage[breaklinks=true,bookmarks=false]{hyperref}

\iccvfinalcopy 

\def\iccvPaperID{1943} 

\ificcvfinal\fi

\newcommand{\Figref}[1]{Fig.~\ref{#1}}
\newcommand{\figref}[1]{Fig.~\ref{#1}}

\newcommand{\Tabref}[1]{Table~\ref{#1}}
\newcommand{\tabref}[1]{Table~\ref{#1}}

\newcommand{\secref}[1]{\S\ref{#1}}

\newcommand{\via}{\emph{via}~}

\definecolor{climate-cyclone}{RGB}{100, 255, 100}
\definecolor{climate-river}{RGB}{255, 50, 40}
\definecolor{climate-background}{RGB}{128,  128,  128} 

\definecolor{2d3ds-unknown}{RGB}{0,0,0}
\definecolor{2d3ds-beam}{RGB}{254,158,137}
\definecolor{2d3ds-board}{RGB}{85,116,127}
\definecolor{2d3ds-bookcase}{RGB}{255,31,33}
\definecolor{2d3ds-ceiling}{RGB}{241,255,82}
\definecolor{2d3ds-chair}{RGB}{0, 18, 141}
\definecolor{2d3ds-clutter}{RGB}{228, 228, 228}
\definecolor{2d3ds-column}{RGB}{89, 173, 163}
\definecolor{2d3ds-door}{RGB}{113, 143, 65}
\definecolor{2d3ds-floor}{RGB}{102, 168, 226}
\definecolor{2d3ds-sofa}{RGB}{100, 22, 116}
\definecolor{2d3ds-table}{RGB}{84, 84, 84}
\definecolor{2d3ds-wall}{RGB}{190, 123, 75}
\definecolor{2d3ds-window}{RGB}{0,255,0}

\definecolor{synthia-misc}{RGB}{0,0,0}
\definecolor{synthia-sky}{RGB}{128,128,128}
\definecolor{synthia-building}{RGB}{128,0,0}
\definecolor{synthia-road}{RGB}{128, 64, 128}
\definecolor{synthia-sidewalk}{RGB}{0,0,192}
\definecolor{synthia-fence}{RGB}{64,64,128}
\definecolor{synthia-vegetation}{RGB}{128,128,0}
\definecolor{synthia-pole}{RGB}{192, 192, 128}
\definecolor{synthia-car}{RGB}{64, 0, 128}
\definecolor{synthia-sign}{RGB}{192, 128, 128}
\definecolor{synthia-pedestrian}{RGB}{64,64,0}
\definecolor{synthia-cyclist}{RGB}{0, 128, 192}
\definecolor{synthia-lanemarking}{RGB}{0, 175,0}
\definecolor{synthia-invalid}{RGB}{255,255,255}
\definecolor{synthia-invalid2}{RGB}{240,240,240}

\begin{document}

\title{Orientation-aware Semantic Segmentation on Icosahedron Spheres}

\author[1]{Chao Zhang\thanks{Equal contribution: \{chao.zhang, stephan.liwicki\}@crl.toshiba.co.uk}}
\newcommand\CoAuthorMark{\footnotemark[\arabic{footnote}]}
\author[1]{Stephan Liwicki\protect\CoAuthorMark}
\author[2]{William Smith}
\author[1,3]{Roberto Cipolla}
\affil[1]{Toshiba Research Europe Limited, Cambridge, United Kingdom\authorcr}
\affil[2]{University of York, United Kingdom}
\affil[3]{University of Cambridge, United Kingdom}

\maketitle
\ificcvfinal\thispagestyle{empty}\fi
\begin{abstract}
    We address semantic segmentation on omnidirectional images, to leverage a holistic understanding of the surrounding scene for applications like autonomous driving systems. For the spherical domain, several methods recently adopt an icosahedron mesh, but systems are typically rotation invariant or require significant memory and parameters, thus enabling execution only at very low resolutions. In our work, we propose an orientation-aware CNN framework for the icosahedron mesh. Our representation allows for fast network operations, as our design simplifies to standard network operations of classical CNNs, but under consideration of north-aligned kernel convolutions for features on the sphere. We implement our representation and demonstrate its memory efficiency up-to a level-8 resolution mesh (equivalent to $640 \times 1024$ equirectangular images). Finally, since our kernels operate on the tangent of the sphere, standard feature weights, pretrained on perspective data, can be directly transferred with only small need for weight refinement. In our evaluation our orientation-aware CNN becomes a new state of the art for the recent 2D3DS dataset, and our Omni-SYNTHIA version of SYNTHIA. Rotation invariant classification and segmentation tasks are additionally presented for comparison to prior art. \vspace{-1em}
\end{abstract}


\input{sections/sec1_introduction.tex}

\input{sections/sec2_related.tex}

\input{sections/sec3_method.tex}

\input{sections/sec4_evaluation.tex}

\input{sections/sec5_conclusion.tex}

{\small
\bibliographystyle{ieee}
\bibliography{SCNN}
}

\end{document}


\title{Supplemental Material: \\ Orientation-aware Semantic Segmentation on Icosahedron Spheres}

\author{First Author\\
Institution1\\
Institution1 address\\
{\tt\small firstauthor@i1.org}
\and
Second Author\\
Institution2\\
First line of institution2 address\\
{\tt\small secondauthor@i2.org}
}

\maketitle

\section{CNN Operation on the Icosahedron Mesh}

We include the pseudo code of our main CNN operators, applied to the icosahedron mesh components, denoted $\{C_i\}_{i=0}^4$. Note, many operations will be a direct result of a combination of these operators (\ie Pyramid Pooling Layers \cite{zhao2017pspnet}). 

First we detail padding in Algorithm~\ref{alg:pad}. Our orientation-aware hexagonal convolutions with arc-based interpolations for north-alignment are given in Algorithm~\ref{alg:convs}. Algorithm~\ref{alg:pool} and \ref{alg:upsample} present pooling and up-sampling respectively. We emphasize, convolutions with kernel size 1, batch normalization, non-linearities and biases are directly computed on the spherical components without padding, through standard unchanged CNN operators. 
\begin{algorithm}[b]
\small
\caption{Padding \& WestPadding (top \& left only)}\label{alg:pad}
\SetAlgoLined
\KwResult{Given sphere components $\{C_i\}_{i=0}^4$ of height $2W$ and width $W$ compute padded $\{P_i\}_{i=0}^4$ }
 \For(\tcp*[f]{pad each component}){$i \leftarrow \{0, \dots, 4\}$}{
  $C_w \leftarrow C_{\mathtt{mod}(i-1,5)}$\tcp*{west neighbor}
  $T \leftarrow \left[\begin{array}{c c c c} 
  C_w(W,W) & \mbox{to} & C_w(1,W) & 0\end{array}\right]$\;
  $L \leftarrow \left[\begin{array}{c}
  \left[\begin{array}{c c c}
  C_w(W+1,W) & \mbox{to} & C_w(2W,W)
  \end{array}\right]^\mathtt{T} \\
  \left[\begin{array}{c c c}
  C_w(2W,W - 1) & \mbox{to} & C_w(2W,1)\end{array}\right]^\mathtt{T} \\
  0 \end{array}\right]$\;
  $P_i \leftarrow \left[ \begin{array}{c}
  T \\
  \left[\begin{array}{c c} L & C_i
  \end{array}\right]
  \end{array} \right]$\tcp*{top \& left}
  \If{pad all sides}{
        $C_e \leftarrow C_{\mathtt{mod}(i+1,5)}$\tcp*{east neighbor}
        $B \leftarrow \left[ \begin{array}{c c c c}
        0 & C_e(2W,1) & \mbox{to} & C_e(W+1,1) \end{array}\right]$\;
        $R \leftarrow \left[ \begin{array}{c}
        0 \\
        \left[\begin{array}{c c c}
        C_e(1,W) & \mbox{to} & C_e(1,1)\end{array}\right]^\mathtt{T}\\
        \left[\begin{array}{c c c}
        C_e(1,1) & \mbox{to} & C_e(W+1,1)\end{array}\right]^\mathtt{T}\end{array}\right]$\;
        $P_i \leftarrow \left[ \begin{array}{c c} \left[ \begin{array}{c}
                      P_i \\
                      B \end{array} \right] & R
                      \end{array}\right]$\tcp*{bottom \& right}
   }
 }
\end{algorithm}
\begin{algorithm}[h]
\small
\caption{Hexagonal Convolution (HexConv)}\label{alg:convs}
\SetAlgoLined
\KwResult{Given components $\{C_i\}_{i=0}^4$ and precomputed interpolation weights $\{A_i\}_{i=0}^4$ get filter results $\{F_i\}_{i=0}^4$ of same size. }
 $\{C_i\}_{i=0}^4 \leftarrow \mathrm{Padding}(\{C_i\}_{i=0}^4)$\tcp*{Alg.~\ref{alg:pad}}
 $\mathbf{W}_1 \leftarrow \left[\begin{array} {c c c}
 w_2 & w_1 & 0 \\
 w_3 & w_7 & w_6 \\
 0  & w_4 & w_5 \end{array}\right]$\tcp*{Hexagon filter}
 $\mathbf{W}_2 \leftarrow \left[\begin{array} {c c c}
 w_3 & w_2 & 0 \\
 w_4 & w_7 & w_1 \\
 0  & w_5 & w_6 \end{array}\right]$\tcp*{Shift weights}
 \For{$i \leftarrow \{0, \dots, 4\}$}{
    $F_i^{1} \leftarrow \mathrm{conv2d}(C_i,\mathbf{W}_1)$\tcp*{standard 2D conv}
    $F_i^{2} \leftarrow \mathrm{conv2d}(C_i,\mathbf{W}_2)$\;
    \tcp{Element-wise Interpolation}
    $F_i \leftarrow A_i \otimes F_i^{1} + (1 - A_i) \otimes F_i^{2}$
 }
\end{algorithm}
\begin{algorithm}[h]
\small
\caption{Pooling on Sphere}\label{alg:pool}
\SetAlgoLined
\KwResult{Given components $\{C_i\}_{i=0}^4$ get pooling $\{F_i\}_{i=0}^4$. }
 $\{C_i\}_{i=0}^4 \leftarrow \mathrm{WestPadding}(\{C_i\}_{i=0}^4)$\tcp*{Alg.~\ref{alg:pad}}
 \For{$i \leftarrow \{0, \dots, 4\}$}{
    $F_i \leftarrow \mathrm{pooling}(C_i)$\tcp*{stride 2 pooling}
 }
\end{algorithm}
\begin{algorithm}[h]
\small
\caption{Bi-linear Up-sampling on Sphere}\label{alg:upsample}
\SetAlgoLined
\KwResult{Given components $\{C_i\}_{i=0}^4$ get bi-linear up-sampling $\{F_i\}_{i=0}^4$. }
 $\{C_i\}_{i=0}^4 \leftarrow \mathrm{WestPadding}(\{C_i\}_{i=0}^4)$\tcp*{Alg.~\ref{alg:pad}}
 \For{$i \leftarrow \{0, \dots, 4\}$}{
    $F_i \leftarrow \mathrm{upsample}(C_i)$\tcp*{$2\times$ up-sampling}
    Cut 1 pixel width from all sides of $F_i$\;
 }
\end{algorithm}

\section{Evaluation Details}
In this section, we include details of network architectures and parameters used in our experiments. 

\begin{table}[h]
    \centering
    \begin{tabular}{cccccc}
         Level & a  &   Block       &   b   &    c   &   s    \\ \hline
         4 & 1      &   HexConv     &  --   &  16    &   1       \\
         4 & 16     &   ResBlock    & 64    & 64     &   2       \\
         3 & 64     &   ResBlock    & 256   & 256    & 2         \\
         2 & 256    &   MaxPool     & --    & --     & 1        \\
         -- & 256   &   Dense       & --    & 10     & 1  \\ \hline
    \end{tabular}
    \caption{HexRUNet-C architecture used in Omni-MNIST experiments. $a,b,c$ stands for input channels, bottleneck channels, and output channels. $s$ stands for strides: 2 means downsampling. }
    \label{tab:mnist_network_jiang}
\end{table}
\begin{table}[t]
    \centering
    \small
    \begin{tabular}{cc}
         Branch 1 & Branch 2  \\ \hline
         -- & Conv2D 1/1, (a, b),pool,BN,f \\
         -- & HexConv, (b, b),BN,f \\
         Conv2D 1/1, (a, c),pool,BN & Conv2D 1/1, (b, c),BN \\ \hline
         \multicolumn{2}{c}{add, f} \\ \hline
    \end{tabular}
    \caption{Residual block (ResBlock), where $a, b, c$ stands for input channels, bottleneck channels and output channels. BN is short for Batch Normalization, and f stands for Rectified Linear Unit activation function (ReLU). }
    \label{tab:res_net}
\end{table}

\subsection{Omni-MNIST}
The input signal for this experiment is on a level-4 mesh, we use max pooling before the final dense layer rather than average pooling used in \cite{jiang2019spherical}. We train our network HexRUNet\=/C with a batch size of 15, initial learning rate of 0.001, and use the Adam optimizer. We use the cross-entropy loss for the digits classification task. The network structure is illustrated in Table \ref{tab:mnist_network_jiang}. The residual block is used across this and other networks, and is shown in Table~\ref{tab:res_net}. Total number of parameters of HexRUNet-C is 74,730.

\subsection{Climate Pattern}
The input signal for this experiment is on a level-5 mesh, the number of input channels is 16.  We use the same architecture as the semantic segmentation task in \S\ref{subsec:2d3ds} (Table~\ref{tab:seg_network_jiang}). We have included two variants using 8 or 32 as the feature maps in the first HexConv operation, called HexRUNet-8 and HexRUNet-32. We train our network with a batch size  60, initial learning rate of 0.001 with Adam optimizer. We train using weighted cross-entropy loss, due to the unbalanced classes distributions. Total number of parameters is 7,543,331 for HexRUNet-32 and 476,747 for HexRUNet\=/8. UGSCNN \cite{jiang2019spherical} uses 8 initial feature maps with a total of 328,339 parameters.

\begin{table}[t]
    \centering
    \begin{tabular}{cccccc}
         Level & a  & Block & b  &  c & s \\ \hline
         5      & 4       & HexConv  & --  & 16  & 1 \\  
         5      & 16      & ResBlock & 16  & 32  & 2 \\
         4      & 32      & ResBlock & 32  & 64 & 2 \\
         3      & 64     & ResBlock & 64 & 128 & 2 \\
         2      & 128     & ResBlock & 128 & 256 & 2 \\
         1      & 256     & ResBlock & 256 & 256 & 2 \\
         0      & 256     & HexConv\textsuperscript{T} & -- & 256 & 0.5 \\ \hline
         1      & 256x2     & ResBlock & 128 & 128 & 0.5 \\
         2      & 128x2   & ResBlock  & 64 & 64 & 0.5 \\
         3      & 64x2   & ResBlock  & 32 & 32  & 0.5 \\
         4      & 32x2    & ResBlock  & 16 & 16  & 0.5 \\
         5      & 16x2    & ResBlock  & 16 & 16  & 1   \\ \hline
         5      & 16      & HexConv   & -- & 13  & 1   \\ \hline
    \end{tabular}
    \caption{HexRUNet architecture used in 2D3DS semantic segmentation experiments. $a,b,c$ stands for input channels, bottleneck channels, and output channels. $s$ stands for strides. When $s=2$, down-sampling is performed, and when $s=0.5$, up-sampling is done (using up-sampling and point-wise HexConv).}
    \label{tab:seg_network_jiang}
\end{table}

\subsection{2D3DS}\label{subsec:2d3ds}
The input signal for this experiment is on a level-5 mesh, and the number of input channels is 4 for RGB and Depth. The network structure is illustrated in Table \ref{tab:seg_network_jiang}. The network contains two parts: encoder layers and decoder layers. At level-0, we apply upsampling and a point-wise HexConv operation to increase the resolution to level-1 (denoted HexConv\textsuperscript{T}). In the subsequent layers, the input will be concatenated with the output from previous layers at corresponding levels. This can be seen for rows in which the input size is doubled.  

We train our network with a batch size 32, initial learning rate of 0.001 with Adam optimizer, up to 500 epochs. In contrast to UGSCNN \cite{jiang2019spherical} which uses 32 feature maps and 5,180,239 parameters, we employ 16 feature maps for the first layer, resulting in 1,585,885 parameters to ensure a competitive comparison between the frameworks. Following \cite{jiang2019spherical}, class-wise weighted cross-entropy loss is used to balance the class examples. Note that the number of output channels is 13 rather than 15, since the 2D3DS dataset has two invalid classes (``unknown'' and ``invalid''), which are not evaluated during validation.

\subsection{Omni-SYNTHIA}
We create our own Omni-SYNTHIA dataset from SYNTHIA data \cite{ros2016synthia}. In the original SYNTHIA data, synthetic views are captured with a stereo set-up consisting of 2 clusters of 4 cameras. For Omni-SYNTHIA we use the left-stereo cluster which captures 4 viewpoints with a common camera center. The views capture $90^\circ$ intervals with a filed of view of $100^\circ$ each. We use the visual overlap to create an omnidirectional view to our needs (\figref{fig:omni_synthia}). Since perspective images are of resolution $760 \times 1280$ the final equirectangular RGB images are set to $2096 \times 4192$. In particular, we keep height/width ratio 1 to 2, and compute the overlap between adjacent viewpoints to find the needed equirectangular resolution. 

Multiple sequences are acquired simulating different cities of four seasons with drastic change of appearance. Ground-truth includes pixel-wise semantic labels of 14 classes (including ``invalid''). In our experiments, the five ``SUMMER" sequences are chosen to make our omnidirectional dataset. Specifically, sequences simulating New York-like (1 and 2) and Highway-like (5 and 6)  scenes are used as training set, while European-like sequence (4) is employed for validation. For each sequence, we choose every second frame. In total, 2,269 equirectangular RGB images are generated (1818 for training, 451 for testing). Depth maps are not used in this experiment.
\begin{figure}[t]
    \centering
    \includegraphics[trim={0cm 20cm 0cm 20cm},clip,width=\linewidth]{dataset_syn/supp/000026-rgb.png}
    \includegraphics[trim={0cm 20cm 0cm 20cm},clip,width=\linewidth]{dataset_syn/supp/000026-label.png}
    \caption{An example of our Omni-SYNTHIA dataset images (top) and labels (bottom). (Top and bottom parts of the images are cropped only for visualization.)}
    \label{fig:omni_synthia}
\end{figure}

\begin{table}
    \centering
    \begin{tabular}{ccc}
         Input & Operator & Output  \\ \hline
         a & HexConv,BN,f & c \\
         c & HexConv,BN,f & c \\
         c & Pool & c \\ \hline
    \end{tabular}
    \caption{U-Net encoder block (Encoder), where $a, c$ stands for input channels and output channels. BN is short for Batch Normalization, and f stands for Rectified Linear Unit activation function (ReLU). }
    \label{tab:unet_down}
\end{table}
\begin{table}
    \centering
    \begin{tabular}{ccc}
         Input & Operator  & Output  \\ \hline
         a & HexConv,BN,f & b \\
         b & HexConv,BN,f & b \\
         b & Up & b \\
         b & Conv2D 1/1, BN, f & c \\\hline
    \end{tabular}
    \caption{U-Net decoder block (Decoder), where $a, b, c$ stands for input channels, middle channels and  output channels. BN is short for Batch Normalization, and f stands for Rectified Linear Unit activation function (ReLU). }
    \label{tab:unet_up}
\end{table}
\begin{table}[]
    \centering
    \begin{tabular}{cccccc}
         Level & a  & Block & b  &  c & s \\ \hline
         6      & 3       & Encoder  & --  & 32  & 2 \\  
         5      & 32      & Encoder & --  & 64  & 2 \\
         4      & 64      & Encoder & --  & 128 & 2 \\
         3      & 128     & Encoder & -- & 256 & 2 \\ \hline
         2      & 256     & Decoder & 512 & 256 & 0.5 \\
         3      & 256x2     & Decoder & 256 & 128 & 0.5 \\
         4      & 128x2     & Decoder & 128  & 64  & 0.5 \\
         5      & 64x2     & Decoder & 64 & 32 & 0.5 \\ \hline
         6      & 32x2   & HexConv,BN,f  & -- & 32  & 1 \\
         6      & 32     & HexConv,BN,f  & -- & 32  & 1 \\
         6      & 32     & HexConv  & -- & 13  & 1  \\\hline
    \end{tabular}
    \caption{HexUNet architecture used in Omni-SYNTHIA semantic segmentation experiments. $a,b,c$ stands for input channels, bottleneck channels, and output channels. $s$ stands for strides. When $s=2$, downsampling is performed, and when $s=0.5$, up-sampling is applied using bi-linear up-sampling and a point-wise convolution.}
    \label{tab:seg_network_unet}
\end{table}

In this experiment, we use the standard U-Net architecture \cite{ronneberger2015u} to facilitate weight transfer from planar U-Net. We call this ``HexUNet'', and the architecture is illustrated in \Tabref{tab:seg_network_unet}. \Tabref{tab:unet_down} and \ref{tab:unet_up} show the detailed encoder and decoder block. Our model has a total of 7,245,101 parameters. Batch size 32, 8 and 2 are used for resolution level 6, 7 and 8 respectively to ensure memory fit on our GPU.

\paragraph{Comparision with state of the art} Spherical input at level-6 is the maximum resolution we could fit in GPU using the provided implementation of \cite{jiang2019spherical}, so we choose to compare our method to UGSCNN using data sampled at level-6 mesh. Planar U-Net \cite{ronneberger2015u} using original perspective images is also evaluated. Images are sub-sampled to match the icosahedron resolution. 

Specifically, we count the number of vertices on the icosahedron mesh that fall onto the image region. We then set the image resolution to be approximately equivalent to this number of vertices, resulting in image resolution $48\times 80$ for level-6, $96\times160$ for level-7 and $192\times320$ for level-8 meshes. To compare network efficiency in terms of training time, we show the average training time on the Omni-SYNTHIA dataset. Evaluations are performed on a single Nvidia 1080Ti GPU with 11 Gb memory. Average training times are obtained by averaging the first 10 epochs. 

\paragraph{Evaluation of Perspective Weights Transfer} 
Using an orientation-aware hexagonal convolution kernel, our method allows direct weights transfer from perspective networks. Initialized with the learned filters of U-Net, we report the results as HexUNet-T in Table \ref{tab:synthia_transfer}. To show the effectiveness of direct weights transfer, we limit weight refinement to up to 10 epochs. Results after just one retraining epoch are  shown in Table~\ref{tab:synthia_transfer}. Our transfer variant achieves competitive results at resolution $r=8$, comparing to source network UNet and our spherical HexUNet trained on up to 500 epochs.

\begin{table}[t]
    \centering
    \scalebox{1}{\begin{tabular}{l | c | c | c}
                        Method    & $r = 6$ & $r = 7$ & $r = 8$ \\ \hline
         UNet (Perspective)& 38.8 & 44.6 & 43.8 \\
         \textbf{HexUNet-T (1 epoch)}  & 29.4  & 30.3 & 35.9 \\
         \textbf{HexUNet-T (10 epochs)} & 36.7  & 38.0 & 45.3 \\
            HexUNet (500 epochs) & 43.6  & 48.3 & 47.1\\ \hline
    \end{tabular}}\vspace{0.5em}
    \caption{Comparison of perspective weights transfer on Omni-SYNTHIA.}
    \label{tab:synthia_transfer}
\end{table}
\begin{figure*}[t]
    \centering
    \subfloat{\begin{minipage}{0.8em}\vspace{-10.3em}\rotatebox{90}{RGB}\end{minipage}}\hfill%
    \includegraphics[trim={1cm 1cm 1cm 1cm},clip, width=0.19\linewidth]{results_2d3ds_supp/camera_e46d5d711e0148f2a141850c5aefadb1_auditorium_2_frame_equirectangular_domain_rgb_data-3.png}\hfill%
    \includegraphics[trim={1cm 1cm 1cm 1cm},clip, width=0.19\linewidth]{results_2d3ds_supp/camera_1ce1d8a48ce249719c2e296a32f7dc49_office_13_frame_equirectangular_domain_rgb_data-5.png}\hfill%
    \includegraphics[trim={1cm 1cm 1cm 1cm},clip, width=0.19\linewidth]{results_2d3ds_supp/camera_09e99df100cb4d59bdaabe5aa8f87ecc_office_42_frame_equirectangular_domain_rgb_data-5.png} \hfill%
    \includegraphics[trim={1cm 1cm 1cm 1cm},clip, width=0.19\linewidth]{results_2d3ds_supp/camera_ce96c72f6dcb4d88a0cfb8d226387421_lobby_1_frame_equirectangular_domain_rgb_data-7.png}\hfill%
    \includegraphics[trim={1cm 1cm 1cm 1cm},clip, width=0.19\linewidth]{results_2d3ds_supp/camera_6a82418b24c54b56b7b288f71e350b57_office_18_frame_equirectangular_domain_rgb_data-3.png} \\
      
    \subfloat{\begin{minipage}{0.8em}\vspace{-10.3em}\rotatebox{90}{GT}\end{minipage}}\hfill%
    \includegraphics[trim={1cm 1cm 1cm 1cm},clip, width=0.19\linewidth]{results_2d3ds_supp/camera_e46d5d711e0148f2a141850c5aefadb1_auditorium_2_frame_equirectangular_domain_rgb_gt-3.png}\hfill%
    \includegraphics[trim={1cm 1cm 1cm 1cm},clip, width=0.19\linewidth]{results_2d3ds_supp/camera_1ce1d8a48ce249719c2e296a32f7dc49_office_13_frame_equirectangular_domain_rgb_gt-5.png}\hfill%
    \includegraphics[trim={1cm 1cm 1cm 1cm},clip, width=0.19\linewidth]{results_2d3ds_supp/camera_09e99df100cb4d59bdaabe5aa8f87ecc_office_42_frame_equirectangular_domain_rgb_gt-5.png} \hfill%
    \includegraphics[trim={1cm 1cm 1cm 1cm},clip, width=0.19\linewidth]{results_2d3ds_supp/camera_ce96c72f6dcb4d88a0cfb8d226387421_lobby_1_frame_equirectangular_domain_rgb_gt-7.png}\hfill%
    \includegraphics[trim={1cm 1cm 1cm 1cm},clip, width=0.19\linewidth]{results_2d3ds_supp/camera_6a82418b24c54b56b7b288f71e350b57_office_18_frame_equirectangular_domain_rgb_gt-3.png}\\
      
    \subfloat{\begin{minipage}{0.8em}\vspace{-10.3em}\rotatebox{90}{UGSCNN\cite{jiang2019spherical}}\end{minipage}}\hfill%
    \includegraphics[trim={1cm 1cm 1cm 1cm},clip, width=0.19\linewidth]{results_2d3ds_supp/camera_e46d5d711e0148f2a141850c5aefadb1_auditorium_2_frame_equirectangular_domain_rgb_jiang-3.png}\hfill%
    \includegraphics[trim={1cm 1cm 1cm 1cm},clip, width=0.19\linewidth]{results_2d3ds_supp/camera_1ce1d8a48ce249719c2e296a32f7dc49_office_13_frame_equirectangular_domain_rgb_jiang-5.png}\hfill%
    \includegraphics[trim={1cm 1cm 1cm 1cm},clip, width=0.19\linewidth]{results_2d3ds_supp/camera_09e99df100cb4d59bdaabe5aa8f87ecc_office_42_frame_equirectangular_domain_rgb_jiang-5.png} \hfill%
    \includegraphics[trim={1cm 1cm 1cm 1cm},clip, width=0.19\linewidth]{results_2d3ds_supp/camera_ce96c72f6dcb4d88a0cfb8d226387421_lobby_1_frame_equirectangular_domain_rgb_jiang-7.png}\hfill%
    \includegraphics[trim={1cm 1cm 1cm 1cm},clip, width=0.19\linewidth]{results_2d3ds_supp/camera_6a82418b24c54b56b7b288f71e350b57_office_18_frame_equirectangular_domain_rgb_jiang-3.png}\\
      
    \subfloat{\begin{minipage}{0.8em}\vspace{-10.3em}\rotatebox{90}{HexRUNet}\end{minipage}}\hfill%
    \subfloat[a]{\includegraphics[trim={1cm 1cm 1cm 1cm},clip, width=0.19\linewidth]{results_2d3ds_supp/camera_e46d5d711e0148f2a141850c5aefadb1_auditorium_2_frame_equirectangular_domain_rgb_hex-3.png}}\hfill%
    \subfloat[b]{\includegraphics[trim={1cm 1cm 1cm 1cm},clip, width=0.19\linewidth]{results_2d3ds_supp/camera_1ce1d8a48ce249719c2e296a32f7dc49_office_13_frame_equirectangular_domain_rgb_hex-5.png}}\hfill%
    \subfloat[c]{\includegraphics[trim={1cm 1cm 1cm 1cm},clip, width=0.19\linewidth]{results_2d3ds_supp/camera_09e99df100cb4d59bdaabe5aa8f87ecc_office_42_frame_equirectangular_domain_rgb_hex-5.png}} \hfill%
    \subfloat[d]{\includegraphics[trim={1cm 1cm 1cm 1cm},clip, width=0.19\linewidth]{results_2d3ds_supp/camera_ce96c72f6dcb4d88a0cfb8d226387421_lobby_1_frame_equirectangular_domain_rgb_hex-7.png}}\hfill%
    \subfloat[e]{\includegraphics[trim={1cm 1cm 1cm 1cm},clip, width=0.19\linewidth]{results_2d3ds_supp/camera_6a82418b24c54b56b7b288f71e350b57_office_18_frame_equirectangular_domain_rgb_hex-3.png}}\\

    \scalebox{0.9}{\begin{tabular}{l l l l l l l}
    \fcolorbox{black}{2d3ds-beam}{\rule{0pt}{6pt}\rule{6pt}{0pt}} beam &
    \fcolorbox{black}{2d3ds-board}{\rule{0pt}{6pt}\rule{6pt}{0pt}} board &
    \fcolorbox{black}{2d3ds-bookcase}{\rule{0pt}{6pt}\rule{6pt}{0pt}} bookcase &
    \fcolorbox{black}{2d3ds-ceiling}{\rule{0pt}{6pt}\rule{6pt}{0pt}} ceiling &
    \fcolorbox{black}{2d3ds-chair}{\rule{0pt}{6pt}\rule{6pt}{0pt}} chair &
    \fcolorbox{black}{2d3ds-clutter}{\rule{0pt}{6pt}\rule{6pt}{0pt}} clutter &
    \fcolorbox{black}{2d3ds-column}{\rule{0pt}{6pt}\rule{6pt}{0pt}} column  \\
    \fcolorbox{black}{2d3ds-door}{\rule{0pt}{6pt}\rule{6pt}{0pt}} door &
    \fcolorbox{black}{2d3ds-floor}{\rule{0pt}{6pt}\rule{6pt}{0pt}} floor &
    \fcolorbox{black}{2d3ds-sofa}{\rule{0pt}{6pt}\rule{6pt}{0pt}} sofa & 
    \fcolorbox{black}{2d3ds-table}{\rule{0pt}{6pt}\rule{6pt}{0pt}} table &
    \fcolorbox{black}{2d3ds-wall}{\rule{0pt}{6pt}\rule{6pt}{0pt}} wall & 
    \fcolorbox{black}{2d3ds-window}{\rule{0pt}{6pt}\rule{6pt}{0pt}} window &
    \fcolorbox{black}{2d3ds-unknown}{\rule{0pt}{6pt}\rule{6pt}{0pt}} unknown
    \end{tabular}}\vspace{0.5em}
    \caption{Qualitative segmentation results and failed cases on 2D3DS dataset.}
    \label{fig:2d3ds_comparison}
\end{figure*}
\section{Additional Results}
We show additional semantic segmentation results for semantic segmentation on 2D3DS and Omni-SYNTHIA.

\subsection{2D3DS Results}

We show additional semantic segmentation results in \Figref{fig:2d3ds_comparison}. Cases (a-c) demonstrate examples on which both our proposed method and UGSCNN \cite{jiang2019spherical} fail to predict the correct labels for some objects. The RGB data of case (a) shows bright illuminations which our method wrongly consider as windows. Windows and bookcases are confused in case (b). As shown in case (c), it is challenging to segment the boundary of a bookcase. Our method failed to recognize windows and chairs in case (d). It also poses a hard sample for UGSCNN. Finally, we argue that some of the ground-truth labels are not accurate, for example, while case (e) presents a bookcase some parts of it are labeled as wall. Even though, our method could adequately predict the bookcase. We also believe that with better input resolution, and with better architectures, our proposed method is capable of improving over these cases.  

\subsection{Omni-SYNTHIA}

We conclude with additional results for the Omni-SYNTHIA resolution evaluation. In general, finer segmentation is achieved at higher resolution. Furthermore, as observed in case (a), (c) and (f) segmentation of small objects and boundaries is improved. In (b) an instance of a cyclist is shown. Here the model classifies misc for low resolution, and pedestrian for $r=7$ and $r=8$. An indication that the network architecture is not ideal for capturing context at all resolutions is case (d), where buildings are misclassified as misc.

\begin{figure*}[p]
\centering
\subfloat{\begin{minipage}{1em}\vspace{-6.3em}\rotatebox{90}{RGB}\end{minipage}}\hspace{1em}%
\subfloat{\includegraphics[width=0.3\linewidth]{images/results_synthia/subs/img_000180.png}}\hspace{1em}%
\subfloat{\includegraphics[width=0.3\linewidth]{images/results_synthia/subs/img_000558.png}}\hspace{1em}%
\subfloat{\includegraphics[width=0.3\linewidth]{images/results_synthia/subs/img_000574.png}}\vspace{-2.9em}\\
\subfloat{\begin{minipage}{1em}\vspace{-6.3em}\rotatebox{90}{GT}\end{minipage}}\hspace{1em}%
\subfloat{\includegraphics[width=0.3\linewidth]{images/results_synthia/subs/gt_000180.png}}\hspace{1em}%
\subfloat{\includegraphics[width=0.3\linewidth]{images/results_synthia/subs/gt_000558.png}}\hspace{1em}%
\subfloat{\includegraphics[width=0.3\linewidth]{images/results_synthia/subs/gt_000574.png}}\vspace{-2.9em}\\
\subfloat{\begin{minipage}{1em}\vspace{-6.3em}\rotatebox{90}{$r = 6$}\end{minipage}}\hspace{1em}%
\subfloat{\includegraphics[width=0.3\linewidth]{images/results_synthia/subs/r6_000180.png}}\hspace{1em}%
\subfloat{\includegraphics[width=0.3\linewidth]{images/results_synthia/subs/r6_000558.png}}\hspace{1em}%
\subfloat{\includegraphics[width=0.3\linewidth]{images/results_synthia/subs/r6_000574.png}}\vspace{-2.9em}\\
\subfloat{\begin{minipage}{1em}\vspace{-6.3em}\rotatebox{90}{$r = 7$}\end{minipage}}\hspace{1em}%
\subfloat{\includegraphics[width=0.3\linewidth]{images/results_synthia/subs/r7_000180.png}}\hspace{1em}%
\subfloat{\includegraphics[width=0.3\linewidth]{images/results_synthia/subs/r7_000558.png}}\hspace{1em}%
\subfloat{\includegraphics[width=0.3\linewidth]{images/results_synthia/subs/r7_000574.png}}\vspace{-2.9em}\\
\subfloat{\begin{minipage}{1em}\vspace{-6.3em}\rotatebox{90}{$r = 8$}\end{minipage}}\hspace{1em}%
\subfloat[a]{\includegraphics[width=0.3\linewidth]{images/results_synthia/subs/r8_000180.png}}\hspace{1em}%
\subfloat[b]{\includegraphics[width=0.3\linewidth]{images/results_synthia/subs/r8_000558.png}}\hspace{1em}%
\subfloat[c]{\includegraphics[width=0.3\linewidth]{images/results_synthia/subs/r8_000574.png}}\\
\subfloat{\begin{minipage}{1em}\vspace{-6.3em}\rotatebox{90}{RGB}\end{minipage}}\hspace{1em}%
\subfloat{\includegraphics[width=0.3\linewidth]{images/results_synthia/subs/img_000648.png}}\hspace{1em}%
\subfloat{\includegraphics[width=0.3\linewidth]{images/results_synthia/subs/img_000738.png}}\hspace{1em}%
\subfloat{\includegraphics[width=0.3\linewidth]{images/results_synthia/subs/img_000876.png}}\vspace{-2.9em}\\
\subfloat{\begin{minipage}{1em}\vspace{-6.3em}\rotatebox{90}{GT}\end{minipage}}\hspace{1em}%
\subfloat{\includegraphics[width=0.3\linewidth]{images/results_synthia/subs/gt_000648.png}}\hspace{1em}%
\subfloat{\includegraphics[width=0.3\linewidth]{images/results_synthia/subs/gt_000738.png}}\hspace{1em}%
\subfloat{\includegraphics[width=0.3\linewidth]{images/results_synthia/subs/gt_000876.png}}\vspace{-2.9em}\\
\subfloat{\begin{minipage}{1em}\vspace{-6.3em}\rotatebox{90}{$r = 6$}\end{minipage}}\hspace{1em}%
\subfloat{\includegraphics[width=0.3\linewidth]{images/results_synthia/subs/r6_000648.png}}\hspace{1em}%
\subfloat{\includegraphics[width=0.3\linewidth]{images/results_synthia/subs/r6_000738.png}}\hspace{1em}%
\subfloat{\includegraphics[width=0.3\linewidth]{images/results_synthia/subs/r6_000876.png}}\vspace{-2.9em}\\
\subfloat{\begin{minipage}{1em}\vspace{-6.3em}\rotatebox{90}{$r = 7$}\end{minipage}}\hspace{1em}%
\subfloat{\includegraphics[width=0.3\linewidth]{images/results_synthia/subs/r7_000648.png}}\hspace{1em}%
\subfloat{\includegraphics[width=0.3\linewidth]{images/results_synthia/subs/r7_000738.png}}\hspace{1em}%
\subfloat{\includegraphics[width=0.3\linewidth]{images/results_synthia/subs/r7_000876.png}}\vspace{-2.9em}\\
\subfloat{\begin{minipage}{1em}\vspace{-6.3em}\rotatebox{90}{$r = 8$}\end{minipage}}\hspace{1em}%
\subfloat[d]{\includegraphics[width=0.3\linewidth]{images/results_synthia/subs/r8_000648.png}}\hspace{1em}%
\subfloat[e]{\includegraphics[width=0.3\linewidth]{images/results_synthia/subs/r8_000738.png}}\hspace{1em}%
\subfloat[f]{\includegraphics[width=0.3\linewidth]{images/results_synthia/subs/r8_000876.png}}\\
\scalebox{0.75}{\fcolorbox{black}{synthia-invalid2}{\rule{0pt}{6pt}\rule{6pt}{0pt}} invalid\hspace{0.2em}
        \fcolorbox{black}{synthia-building}{\rule{0pt}{6pt}\rule{6pt}{0pt}} building\hspace{0.2em}
        \fcolorbox{black}{synthia-car}{\rule{0pt}{6pt}\rule{6pt}{0pt}} car\hspace{0.2em}
        \fcolorbox{black}{synthia-cyclist}{\rule{0pt}{6pt}\rule{6pt}{0pt}} cyclist\hspace{0.2em}
        \fcolorbox{black}{synthia-fence}{\rule{0pt}{6pt}\rule{6pt}{0pt}} fence\hspace{0.2em}
        \fcolorbox{black}{synthia-lanemarking}{\rule{0pt}{6pt}\rule{6pt}{0pt}} marking\hspace{0.2em}
        \fcolorbox{black}{synthia-misc}{\rule{0pt}{6pt}\rule{6pt}{0pt}} misc\hspace{0.2em}
        \fcolorbox{black}{synthia-pedestrian}{\rule{0pt}{6pt}\rule{6pt}{0pt}} pedestrian\hspace{0.2em}
        \fcolorbox{black}{synthia-pole}{\rule{0pt}{6pt}\rule{6pt}{0pt}} pole\hspace{0.2em}
        \fcolorbox{black}{synthia-road}{\rule{0pt}{6pt}\rule{6pt}{0pt}} road\hspace{0.2em}
        \fcolorbox{black}{synthia-sidewalk}{\rule{0pt}{6pt}\rule{6pt}{0pt}} sidewalk\hspace{0.2em}
        \fcolorbox{black}{synthia-sign}{\rule{0pt}{6pt}\rule{6pt}{0pt}} sign\hspace{0.2em}
        \fcolorbox{black}{synthia-sky}{\rule{0pt}{6pt}\rule{6pt}{0pt}} sky\hspace{0.2em}
        \fcolorbox{black}{synthia-vegetation}{\rule{0pt}{6pt}\rule{6pt}{0pt}} vegetation}\vspace{0.5em}
\caption{Unfolded visualizations of semantic segmentation results on Omni-SYNTHIA dataset at different resolutions. }
\label{fig:resolutions}
\end{figure*}

{\small
\bibliographystyle{ieee}
\bibliography{SCNN}
}

%% file: sections/sec1_introduction.tex
\section{Introduction}

We address the problem of spherical semantic segmentation on omnidirectional images.
Accurate semantic segmentation is useful for many applications including scene understanding, robotics, and medical image processing. It is also a key component for autonomous driving technology. Deep convolutional neutral networks (CNNs) have pushed the performance on a wide array of high-level tasks, including image classification, object detection and semantic segmentation. In particular, most research on CNNs for semantic segmentation \cite{long2015fully,ronneberger2015u,zhao2017pspnet,chen2018deeplab} thus far has focused on perspective images. 
In our work, we focus on omnidirectional images, as such data provides a holistic understanding of the surrounding scene with a large field of view. The complete receptive field is especially important for autonomous driving systems. Furthermore, recent popularity in omnidirectional capturing devices and the increasing number of datasets with omnidirectional signals make omnidirectional processing very relevant for modern technology.
\begin{figure}
    \centering
    \subfloat{\begin{minipage}{0.9\linewidth}\includegraphics[width=1\linewidth]{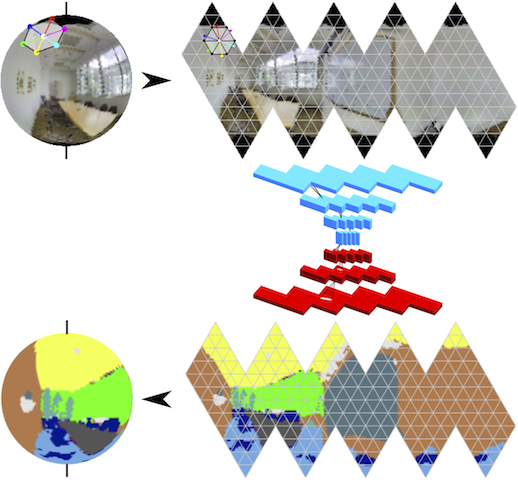}\end{minipage}}
    \subfloat{\hspace{-0.9\linewidth}\begin{minipage}{0.5\linewidth}\scalebox{0.68}{\begin{tabular}{l l}
    \fcolorbox{black}{2d3ds-beam}{\rule{0pt}{6pt}\rule{6pt}{0pt}} beam &
    \fcolorbox{black}{2d3ds-board}{\rule{0pt}{6pt}\rule{6pt}{0pt}} board \\
    \fcolorbox{black}{2d3ds-bookcase}{\rule{0pt}{6pt}\rule{6pt}{0pt}} bookcase &
    \fcolorbox{black}{2d3ds-ceiling}{\rule{0pt}{6pt}\rule{6pt}{0pt}} ceiling \\
    \fcolorbox{black}{2d3ds-chair}{\rule{0pt}{6pt}\rule{6pt}{0pt}} chair &
    \fcolorbox{black}{2d3ds-clutter}{\rule{0pt}{6pt}\rule{6pt}{0pt}} clutter \\
    \fcolorbox{black}{2d3ds-column}{\rule{0pt}{6pt}\rule{6pt}{0pt}} column &
    \fcolorbox{black}{2d3ds-door}{\rule{0pt}{6pt}\rule{6pt}{0pt}} door \\
    \fcolorbox{black}{2d3ds-floor}{\rule{0pt}{6pt}\rule{6pt}{0pt}} floor &
    \fcolorbox{black}{2d3ds-sofa}{\rule{0pt}{6pt}\rule{6pt}{0pt}} sofa \\
    \fcolorbox{black}{2d3ds-table}{\rule{0pt}{6pt}\rule{6pt}{0pt}} table &
    \fcolorbox{black}{2d3ds-wall}{\rule{0pt}{6pt}\rule{6pt}{0pt}} wall \\
    \fcolorbox{black}{2d3ds-window}{\rule{0pt}{6pt}\rule{6pt}{0pt}} window &
    \fcolorbox{black}{2d3ds-unknown}{\rule{0pt}{6pt}\rule{6pt}{0pt}} unknown
        \end{tabular}}\end{minipage}}\vspace{0.1em}
    \caption{Given spherical input, we convert it to an unfolded icosahedron mesh. Hexagonal filters are then applied under consideration of north alignment, as we efficiently interpolate vertices. Our approach is suited to most classical CNN architectures, \eg U-Net \cite{ronneberger2015u}. Since we work with spherical data, final segmentation results provide a holistic labeling of the environment.}
    \label{fig:teaser}
    \vspace{-1em}
\end{figure}

While spherical input could be represented as planar equirectangular images where standard CNNs are directly applied, such choice is inferior due to latitude dependent distortions and boundaries. In \cite{su2017learning} a perspective network is distilled to work on equirectangular input. The main drawback is that weight sharing is only enabled in horizontal direction. Therefore, the model requires more parameters than a perspective one. SphereNet \cite{coors2018spherenet} projects equirectangular input onto a latitude-longitude grid. A constant grid kernel is convolved with each vertex on the sphere by sampling on the tangent plane. However, it is not straightforward to implement pooling and up-sampling for dense prediction tasks.

In 3D shape analysis, one of the challenges in applying CNNs is how to define a natural convolution operator on non-euclidean surfaces. Several works \cite{bronstein2017geometric, monti2017geometric,boscaini2016learning} have focused on networks for manifolds or graphs. Unlike general 3D shapes, omnidirectional images are orientable with the existence of north and south poles. Therefore, the lack for shift-invariance on surfaces or graphs could be overcome with an orientation-aware representation. 

Most recently, several works propose to use an icosahedron mesh as the underlying spherical data representation. The base icosahedron is the most regular polyhedron, consisting of 12 vertices and 20 faces. It also provides a simple way of resolution increase \via subdivision. In \cite{jiang2019spherical}, UGSCNN is proposed to use the linear combinations of differential operators weighted by learnable parameters. Since the operaters are precomputed, the number of parameters is reduced to 4 per kernel. The main issue of this approach, as observed in our experiments, is that it requires a lot of memory for their mesh convolution if the resolution is lifted for better input/output quality. Similar to our method in the use of icosahedron, \cite{cohen2019gauge} proposes a gauge equivariant CNN. Here, filter weights are shared across multiple orientations. While rotation covariance and invariance is essential in applications such as 3D shape classification and climate pattern prediction, it might be undesired in the semantic segmentation task which we consider here. On the contrary, we argue that the orientation information from cameras attached to vehicles or drones is very important a cue and should be exploited. 

Therefore, we propose and investigate a novel framework for the application of CNNs to omnidirectional input, targeting semantic segmentation. We take advantage of both, the icosahedron representation for efficiency and orientation information to improve accuracy in orientation-aware tasks (\figref{fig:teaser}). Our hypothesis is that aligning all learnable filters to the north pole is essential for omnidirectional semantic segmentation. We also argue that high resolution meshes (\ie a level-8 icosahedron mesh) are needed for detailed segmentation. Due to memory restrictions, CNN operations need to be implemented efficiently to reach such high resolution.

In our work, we first map the spherical data to an icosahedron mesh, which we unfold along the equator, similarly to cube maps \cite{monroy2018salnet360,cheng2018cube} and \cite{liu2019deep,cohen2019gauge}. In the icosahedron, vertices have at most 6 neighbors. Therefore, we propose to use a hexagonal filter that is applied to each vertex's neighborhood. After simple manipulation of the unfolded mesh, standard planar CNN operations compute our hexagonal convolutions, pooling and up-sampling layers. Finally we emphasize, since our filters are similar to standard $3\times3$ kernels applied to the tangent of the sphere, weight transfer from pretrained perspective CNNs is possible.

To validate our approach we use the omnidirectional 2D3DS dataset \cite{armeni2017joint} and additionally prepare our Omni-SYNTHIA dataset, which is produced from SYNTHIA data \cite{ros2016synthia}. Qualitative as well as quantitative results demonstrate that our method outperforms previous state-of-the-art approaches in both scenarios. Performance on spherical MNIST classification \cite{cohen2018spherical} and climate pattern segmentation \cite{mudigonda2017segmenting} is also shown in comparison with previous methods in literature. In summary, our contributions are:
\begin{enumerate}
    \item We propose and implement a memory efficient icosahedron-based CNN framework for spherical data.\vspace{-0.5em}
    \item We introduce fast interpolation for orientation-aware filter convolutions on the sphere.\vspace{-0.5em}
    \item We present weight transfer from kernels learned through classical CNNs, applied to perspective data.\vspace{-0.5em}
    \item We evaluate our method on both non-orientation-aware and orientation-aware, publicly available datasets.
\end{enumerate}

%% file: sections/sec2_related.tex
\section{Related Work}

\paragraph{CNNs on Equirectangular Images}
Although classical CNNs are not designed for omnidirectional data, they could still be used for spherical input if the data are converted to equirectangular form. Conversion from spherical coordinates to equirectangular images is a linear one-to-one mapping, but spherical inputs are distorted drastically especially in polar regions. Another artifact is that north and south poles are stretched to lines. Lai \etal \cite{lai2018semantic} apply this method in the application of converting panoramic video to normal perspective.  
Another method along this line is to project spherical data onto multiple faces of a convex polygons, such as a cube. In \cite{monroy2018salnet360}, omnidirectional images are mapped to 6 faces of a cube, and then trained with normal CNNs. However, distortions still exist and discontinuities between faces have to be carefully handled. 

\paragraph{Spherical CNNs}
In order to generalize convolution from planar images to spherical signals, the most natural idea is to replace shifts of the plane by rotations of the sphere. Cohen \etal \cite{cohen2018spherical} propose a spherical CNN which is invariant in the $SO(3)$ group. Esteves \etal \cite{esteves2018learning} use spherical harmonic basis to achieve similar results. Zhou \etal\cite{zhou2017oriented} propose to extend normal CNNs to extract rotation-dependent features by including an additional orientation channel.

\paragraph{CNNs with Deformable Kernels}
Some works \cite{dai2017deformable,jeon2017active} consider adapting the sampling locations of convolutional kernels. Dai \etal \cite{dai2017deformable} propose to learn the deformable convolution which samples the input features through learned offsets. An Active Convolutional Unit is introduced in \cite{jeon2017active} to provide more freedom to a conventional convolution by using position parameters. These methods requires additional model parameters and training steps to learn the sampling locations. In our work, we adapt the kernel shape to fit icosahedron geometry. Unlike deformable methods, our sampling locations can be precomputed and reused without the need of training. 

\paragraph{CNNs with Grid Kernels}
Another line of works aim to adapt the regular grid kernel to work on omnidirectional images. Su and Grauman \cite{su2017learning} propose to process equirectangular images as perspective ones by adapting the weights according to the elevation angles. Weight sharing is only enabled along the longitudes. To reduce the computational cost and degradation in accuracy, a Kernel Transformer Network \cite{su2018kernel} is applied to transfer convolution kernels from perspective images to equirectangular inputs. Coors \etal \cite{coors2018spherenet} present SphereNet to minimize the distortions introduced by applying grid kernels on equirectangular images. Here, a kernel of fixed shape is used to sample on the tangent plane according to the location on the sphere. Wrapping the kernel around the sphere avoids cuts and discontinuities. 

\paragraph{CNNs with Reparameterized Kernels}
For the efficiency of CNNs, several works are proposed to use parameterized convolution kernels. Boscani \etal \cite{boscaini2016learning} introduce oriented anisotropic diffusion kernels to estimate dense shape correspondence. 
Cohen and Welling \cite{cohen2016steerable} employ a linear combination of filters to achieve equivariant convolution filters. In \cite{weiler20183d}, 3D steerable CNNs using linear combination of filter banks are developed. Recently, Jiang \etal \cite{jiang2019spherical} utilized parameterized differential operators as spherical convolution for unstructured grid data. Here, a convolution operation is a linear combination of four differential operators with learnable weights. However, these methods are limited to the chosen kernel types and are not maximally flexible.

\paragraph{CNNs on Icosahedron}
Related to our approach in discrete representation, several works utilize an icosahedron for spherical image representation. As the most uniform and accurate discretization of the sphere, the icosahedron is the regular convex polyhedron with the most faces. A spherical mesh can be generated by progressively subdividing each face into four equal triangles and reprojecting each node to unit length. Lee \etal \cite{lee2018spherephd} is one of the first to suggest the use of icosahedrons for CNNs on omnidirectional images. Here, convolution filters are defined in terms of triangle faces. In \cite{jiang2019spherical}, UGSCNN is proposed to efficiently train a convolutional network with spherical data mapped to an icosahedron mesh. Liu \etal\cite{liu2019deep} uses the icosahedron based spherical grid as the discrete representation of the spherical images and proposes an azimuth-zenith anisotropic CNN for 3D shape analysis. Cohen \etal \cite{cohen2019gauge} employ an icosahedron mesh to present a gauge equivariant CNN. Equivariance is ensured by enforcing filter weight sharing across multiple orientations.

%% file: sections/sec3_method.tex
\section{Proposed Spherical Representation} 
\begin{figure}
\centering
\subfloat[Input sphere]{\includegraphics[height=6.5em]{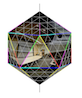}}\hfill
\subfloat[Icosahedron]{\includegraphics[height=6.5em]{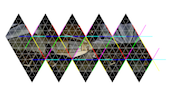}}\hfill
\subfloat[Unfolded representation]{\includegraphics[height=6.5em]{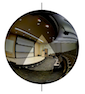}}\\
\subfloat[Image-grid-aligned representation of spherical data]{\includegraphics[width=0.84\linewidth]{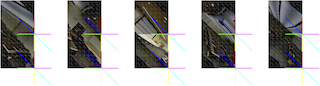}}\\\vspace{0.5em}
\caption{Spherical input data (a) is represented by an icosahedron-based geodesic grid (b). Similar to cubes \cite{monroy2018salnet360,cheng2018cube}, we unfold our mesh (c) and align its 5 components to the standard image grid (d) for efficient computation of convolution, pooling and up-sampling.}\label{fig:icosahedron_flat}
\end{figure}
We represent the spherical input through vertices on an icosahedron mesh (\Figref{fig:icosahedron_flat}). The mapping is based on the vertices' azimuth and zenith angles -- \eg the input color is obtained from an equirectangular input through interpolation. Similar to cube maps \cite{monroy2018salnet360,cheng2018cube}, the icosahedron simplifies the sphere into a set of planar regions. While the cube represents the sphere only with 6 planar regions, the icosahedral representation is the convex geodesic grid with the largest number of regular faces. In total, our gird consists of 20 faces and 12 vertices at the lowest resolution, and $f_r = 20*4^r$ faces and $n_r = 2 + 10 * 4^r$ vertices at resolution level $r \geq 0$. Note, a resolution increase is achieved by subdivision of the triangular faces at $r=0$ into $4^r$ equal regular triangular parts. In the following, we present an efficient orientation-aware implementation of convolutions in \secref{ss:convs}, and our down- and up-sampling techniques in \secref{ss:pool}. Finally, weight transfer from trained kernels of standard perspective CNNs is discussed in \secref{subsec:transfer}.

\subsection{Orientation-aware Convolutions}\label{ss:convs}

If a camera is attached to a vehicle, the orientation and location of objects such as sky, buildings, sidewalks or roads are likely similar across the dataset. Therefore, we believe an orientation-aware system can be beneficial, while tasks with arbitrary rotations may benefit from rotation invariance \cite{cohen2018spherical} or weight sharing across rotated filters \cite{worrall2018cubenet,cohen2019gauge}. 

\paragraph{Efficient Convolutions through Padding}
We first define the north and south pole as any two vertices that have maximum distance on the icosahedron mesh. Similar to \cite{liu2019deep,cohen2019gauge}, the mesh is then converted to a planar representation by unfolding it along the equator (\Figref{fig:icosahedron_flat}). Finally, we split the surface into five components, and align the vertices with a regular image grid through a simple affine transformation.

\begin{figure}
\centering
\subfloat[Convolution]{\includegraphics[height=2.7cm]{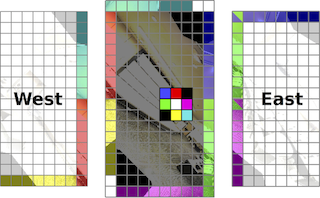}}\hfill
\subfloat[Up-sampling]{\includegraphics[height=2.7cm]{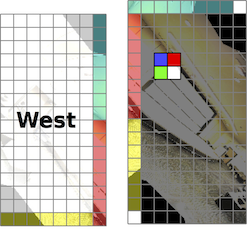}}\vspace{0.5em}
\caption{Convolution with our hexagonal filters (a) and up-sampling (b) reduce to standard CNN operations after padding the sphere component with features from neighboring sphere parts. Pooling is computed with a standard 2x2 kernel with stride 2.}\label{fig:conv}
\end{figure}

Notice, vertices have a neighborhood of either 5 or 6 points. Hence we employ hexagonal filters in our work, instead of regular $3\times3$ kernels. Let us ignore the vertices at the poles (\eg through reasoning of dropout), and adjust the neighborhood cardinality to 6 for all vertices with 5 neighbors through simple repetition. Now, our planar representation of the icosahedron simplifies the convolution with hexagonal filters to standard 2D convolution with a masked kernel, after padding as shown in \Figref{fig:conv}.

\begin{figure}
    \centering
    \subfloat[North-alignment]{\begin{minipage}{0.4\linewidth}\includegraphics[width = 1\linewidth]{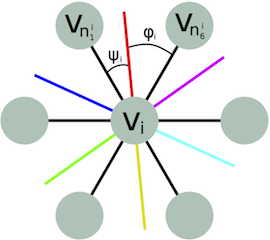}\end{minipage}}\hspace{2em}
    \subfloat[Interpolated filters]{\begin{minipage}{0.454\linewidth}
    \begin{minipage}{0.4\linewidth}\Large$\frac{\phi_i}{\phi_i + \psi_i} *$\end{minipage}
    \begin{minipage}{0.38\linewidth}\includegraphics[width =1\linewidth]{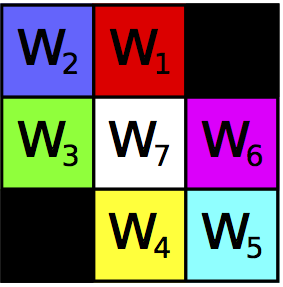}\end{minipage}\\
    \begin{minipage}{0.4\linewidth}\Large$\frac{\psi_i}{\phi_i + \psi_i} *$\end{minipage}
    \begin{minipage}{0.38\linewidth}\includegraphics[width =1\linewidth]{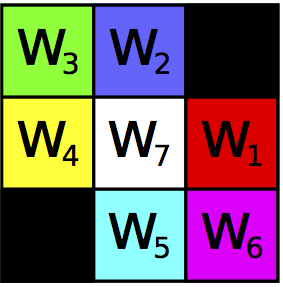}\end{minipage}
    \end{minipage}}\vspace{0.5em}
    \caption{Given arc-based interpolation of the neighborhood for north-alignment (a), our convolution is computed with 2 weighted filters (b). The weights are precomputed for all vertices.}
    \label{fig:interpolation}
\end{figure}

\paragraph{North-alignment through Interpolation} \label{par:north}
In its natural implementation, our filters are aligned to the icosahedron mesh. Consequently, the filter orientation is inconsistent, since the surfaces near the north and south poles are stitched. We reduce the effect of such distortions by aligning filters vertically through interpolation (\Figref{fig:interpolation}).

The na\"ive convolution with weights $\{w_j\}_{j=1}^7$ at vertex $\mathbf{v}_i$ and its neighbors $\{\mathbf{v}_{n_j^i}\}_{j=1}^6$, is computed as $\sum_{j=1}^6 w_j\mathbf{v}_{n_j^i} + w_7\mathbf{v}_i$, where $n_j^i$ holds the neighborhood indices of $\mathbf{v}_i$. Instead, we north-align the neighborhood with interpolations using arc-based weights $\{\theta_j^i\}_{j=1}^6$ as follows:
\begin{align}\label{eq:base_interpolate}
\MoveEqLeft{\sum_{j=2}^6 w_j (\theta_j^i \mathbf{v}_{n_j^i} + (1-\theta_j^i) \mathbf{v}_{n_{j-1}^i} )} \nonumber\\
& + w_1 (\theta_1^i \mathbf{v}_{n_1^i} + (1-\theta_1^i) \mathbf{v}_{n_{6}^i}) + w_7 \mathbf{v}_i.
\end{align}
Since the hexagonal neighborhood is approximately symmetric, we further simplify \eqref{eq:base_interpolate} by introducing a unified weight $\alpha_i$, such that $\{\alpha_i \approx \theta_j^i\}_{j=1}^6$ holds. Hence we write
\begin{align}
\MoveEqLeft \alpha_i \left(\sum_{j=1}^6 w_j \mathbf{v}_{n_j^i} + w_7\mathbf{v}_i \right)\nonumber\\
\MoveEqLeft + (1-\alpha_i) \left( \sum_{j=2}^6 w_{j}\mathbf{v}_{n_{j-1}^i} + w_1\mathbf{v}_{n_6^i} + w_7\mathbf{v}_i \right).
\end{align}
Thus, north-aligned filters can be achieved through 2 standard convolutions, which are then weighted based on the vertices' interpolations $\alpha_i$. 

The arc-interpolation $\alpha_i$ is based on the angle distance between the direction towards the first and sixth neighbors (\ie $\mathbf{v}_{n_1^i}$ and $\mathbf{v}_{n_6^i}$ respectively) and the north-south axis when projected onto the surface of the sphere. In particular, we first find the projective plane of the north-south axis $\mathbf{a} = \left[\begin{array}{c c c}
0 & 1 &0 \end{array}\right]^{\mathtt{T}}$ towards vector $\mathbf{v}_i$ as the plane with normal
$\mathbf{n}_i = \frac{\mathbf{v}_i \times \mathbf{a}}{\left|\mathbf{v}_i \times \mathbf{a}\right|}$.
Since the spherical surface is approximated by the plane of vectors $\mathbf{v}_i - \mathbf{v}_{n_1^i}$ and $\mathbf{v}_i - \mathbf{v}_{n_6^i}$, we only require the angles between these vectors and the plane given by $\mathbf{n}_i$, to find interpolation $\alpha_i = \frac{\phi_i}{\phi_i + \psi_i}$ with
\begin{align}
   \psi_i &= \arccos\frac{(\mathbf{v}_i - \mathbf{v}_{n_1^i})^\mathtt{T}(\mathbf{I} - \mathbf{n}_i\mathbf{n}_i^\mathtt{T})(\mathbf{v}_i - \mathbf{v}_{n_1^i})}{\left|(\mathbf{v}_i - \mathbf{v}_{n_1^i})\right|\left|(\mathbf{I} - \mathbf{n}_i\mathbf{n}_i^\mathtt{T})(\mathbf{v}_i - \mathbf{v}_{n_1^i})\right|} \nonumber\\
   \phi_i &= \arccos\frac{(\mathbf{v}_i - \mathbf{v}_{n_6^i})^\mathtt{T}(\mathbf{I} - \mathbf{n}_i\mathbf{n}_i^\mathtt{T})(\mathbf{v}_i - \mathbf{v}_{n_6^i})}{\left|(\mathbf{v}_i - \mathbf{v}_{n_6^i})\right|\left|(\mathbf{I} - \mathbf{n}_i\mathbf{n}_i^\mathtt{T})(\mathbf{v}_i - \mathbf{v}_{n_6^i})\right|}.
\end{align}

\subsection{Pooling and Up-sampling}\label{ss:pool}

Down-sampling through pooling and bi-linear up-sampling are important building blocks of CNNs, and are frequently employed in the encoder-decoder framework of semantic segmentation (\eg \cite{ronneberger2015u}). Pooling is aimed at summarising the neighborhood of features to introduce robustness towards image translations and omissions. Typically, a very small and non-overlapping neighborhood of $2\times2$ pixels is considered in standard images, to balance detail and redundancy. Bi-linear up-sampling is used in the decoder to increase sub-sampled feature-maps to larger resolutions.

We note, in our icosahedron mesh the number of vertices increases by a factor of $4$ for each resolution (excluding poles). Therefore during down-sampling from resolution $r$ to $r-1$, we summarize a neighborhood of $4$ at $r$ with $1$ vertex at $r-1$. A natural choice is to pool over $\{\mathbf{v}_i, \mathbf{v}_{n_1^i}, \mathbf{v}_{n_2^i} , \mathbf{v}_{n_3^i}\}$ for vertices $\mathbf{v}_i$ that are represented in both resolutions. Thus, we apply a simple standard $2\times2$ strided pooling with kernel $2\times2$ on each icosahedron part.

Analogously, bi-linear up-sampling or transposed convolutions are applied by padding the icosahedron parts at left and top followed by up-sampling by a factor of $2$ in height and width (\Figref{fig:conv}). Due to padding, this results in a $1$-pixel border at each size which we simply remove to provide the expected up-sampling result.

Finally we emphasize, methods like pyramid pooling \cite{zhao2017pspnet} can be computed by combining our pooling and up-sampling techniques.

\subsection{Weight Transfer from Perspective Networks}\label{subsec:transfer}

\begin{figure}
    \centering
    \subfloat{\begin{minipage}{0.2\linewidth}\includegraphics[width=1\linewidth]{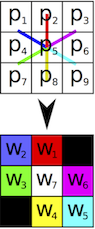}\end{minipage}}\hspace{0.5em}
    \subfloat{\begin{minipage}{0.7\linewidth}\begin{tabular}{l}
     $w_1 = p_2$\\
     $w_2 = \sin\frac{\pi}{3}\frac{p_1 + p_4}{2} + (1-\sin\frac{\pi}{3}) \frac{p_2 + p_5}{2}$\\
     $w_3 = \sin\frac{\pi}{3}\frac{p_4 + p_7}{2} + (1-\sin\frac{\pi}{3}) \frac{p_5 + p_8}{2}$\\
     $w_4 = p_8$ \\
     $w_5 = \sin\frac{\pi}{3}\frac{p_6 + p_9}{2} + (1-\sin\frac{\pi}{3}) \frac{p_5 + p_8}{2}$\\
     $w_6 = \sin\frac{\pi}{3}\frac{p_3 + p_6}{2} + (1-\sin\frac{\pi}{3}) \frac{p_2 + p_5}{2}$\\
     $w_7 = p_5$ \\
\end{tabular}\end{minipage}}\vspace{0.5em}
    \caption{The weights of conventional $3\times3$ kernels trained on perspective data can be transferred to our model \via simple interpolation as our filters operate on the sphere's tangent planes.}
    \label{fig:weight_transfer}
\end{figure}

Similar to SphereNet \cite{coors2018spherenet}, our network applies an oriented filter at the local tangent plane of each vertex on the sphere. Consequently, the transfer of pretrained perspective network weights is naturally possible in our setup. Since we apply hexagonal filters with 7 weights, we interpolate from the standard $3\times3$ kernels as shown in \Figref{fig:weight_transfer}. Specifically, we align north and south of the hexagon with the second and eighth weight of the standard convolution kernel respectively. Bi-linear interpolation provides the remaining values for our filter. After transfer, weight refinement is necessary, but can be computed on a much smaller dataset (as done in \cite{coors2018spherenet}), or reduced learning iterations. Alternatively, but left for future work, it should be possible to learn hexagonal filter weights directly on perspective datasets \cite{sun2016design, hoogeboom2018hexaconv}.

%% file: sections/sec4_evaluation.tex
\section{Evaluation}
The main focus of this paper is omnidirectional semantic segmentation. Both synthetic urban scene and real indoor environments are evaluated. For completeness, we also include our model in comparison with previous state-of-the-art methods on spherical MNIST classification in \secref{subsec:mnist} and a climate pattern prediction task in \secref{subsec:climate}. In \secref{subsec:2d3ds} and \secref{subsec:synthia}, performance on omnidirectional semantic segmentation tasks are summarized and analysed.   

\subsection{Spherical MNIST} \label{subsec:mnist}

We follow \cite{cohen2018spherical} in the preparation of the spherical MNIST dataset, as we prepare non-rotated training and testing (N/N), non-rotated training with rotated testing (N/R) and rotated training and testing (R/R) tasks. 
Both non-rotated and rotated versions are generated using public source code provided by UGSCNN \cite{jiang2019spherical}.\footnote{https://github.com/maxjiang93/ugscnn} 
Training set and test set include 60,000 and 10,000 digits, respectively.
Input signals for this experiment are on a level-4 mesh (\ie $r = 4$). The residual U-Net architecture of \cite{jiang2019spherical}, including the necessary modifications to adapt to the classification task, is used in our experiments. We call this network ``HexRUNet-C''.

\begin{table}[t]
    \centering
    \scalebox{1}{
    \begin{tabular}{c | c c c}
                    Method            & N/N        & N/R         & R/R        \\ \hline
         Spherical CNN \cite{cohen2018spherical}          & 96.\_\_    & \bf{94.\_\_}     & 95.\_\_    \\
         Gauge Net \cite{cohen2019gauge}                 & 99.43      & 69.99       & \bf{99.31}      \\
         UGSCNN \cite{jiang2019spherical}                  & 99.23      & 35.60       & 94.92      \\ \hline
     \textbf{HexRUNet-C}   & \bf{99.45}      & 29.84       & 97.05      \\ \hline
    \end{tabular}}\vspace{0.5em}
    \caption{Spherical MNIST with non-rotated (N) and rotated (R) training and test data. Orientation-aware HexRUNet-C is competitive only when training and test data match (\ie N/N and R/R).}
    \label{tab:mnist_results}
\end{table}
As shown in \Tabref{tab:mnist_results}, our method outperforms previous methods for N/N, achieving 99.45\% accuracy. In R/R, our method performs better than competing Spherical CNN  and UGSCNN. Gauge Net benefits from weight sharing across differently oriented filters, and achieves best accuracy for this task amongst all approaches. Similar to \cite{jiang2019spherical}, our method is orientation-aware by design and thus not rotation-invariant. Therefore, it is expected to not generalize well to randomly rotated test data in the N/R setting, while Spherical CNN performs best in this case.

\subsection{Climate Pattern Segmentation}\label{subsec:climate}

\begin{figure}[t]
    \centering
\subfloat{\includegraphics[width=0.55\linewidth,trim=0 200pt 0pt 200pt,clip=true]{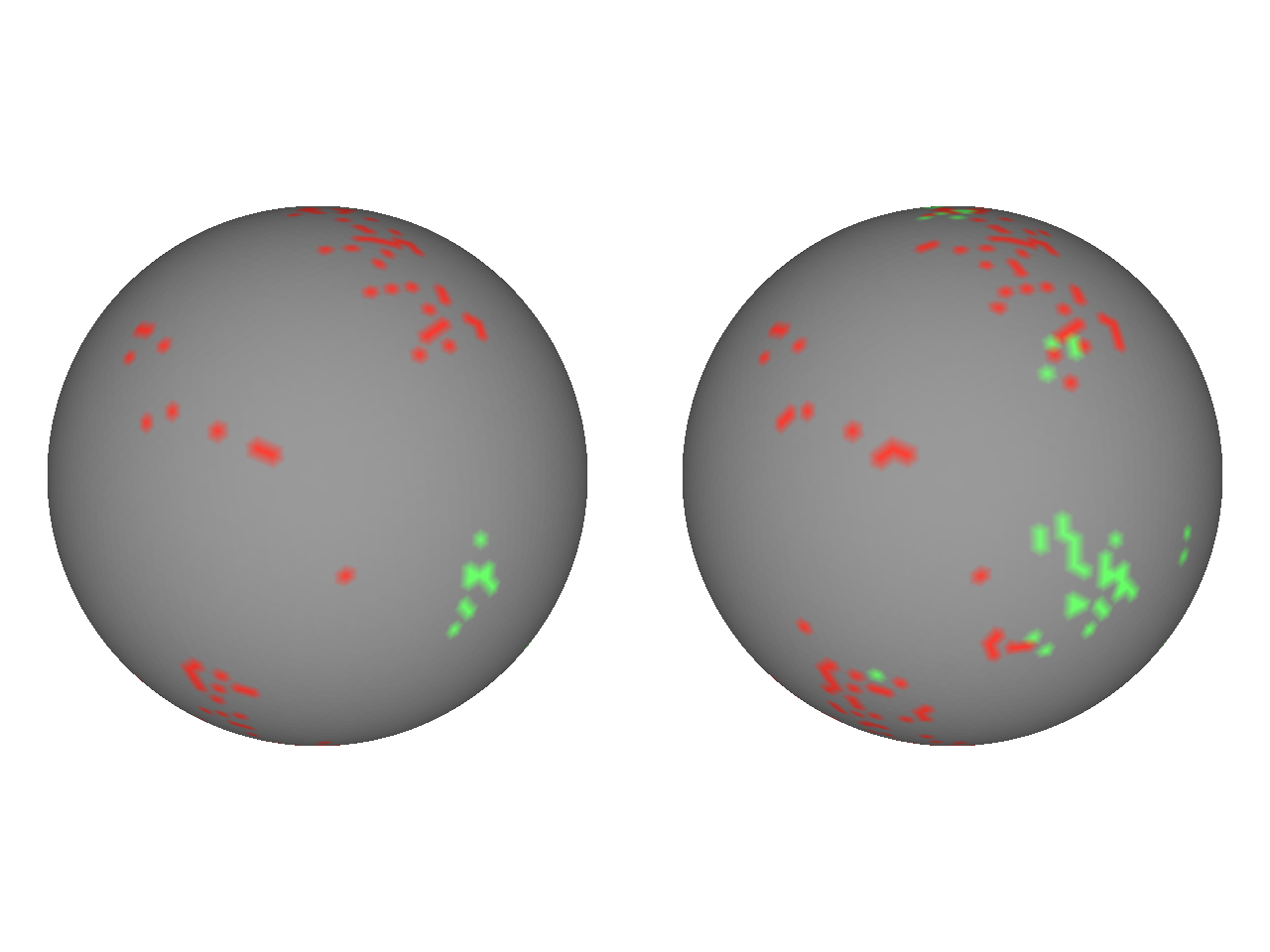}}\hspace{1em}
\subfloat{\begin{minipage}{4em}\vspace{-5em}
\scalebox{0.8}{\fcolorbox{black}{climate-cyclone}{\rule{0pt}{6pt}\rule{6pt}{0pt}} TC} \\ \scalebox{0.8}{\fcolorbox{black}{climate-river}{\rule{0pt}{6pt}\rule{6pt}{0pt}} AR }\\
\scalebox{0.8}{\fcolorbox{black}{climate-background}{\rule{0pt}{6pt}\rule{6pt}{0pt}} BG }
\end{minipage}}
\caption{Semantic segmentation results of HexRUNet-32 on climate pattern (right) in comparison to ground truth (left).}    \label{fig:climate_comparison}
\end{figure}

We further evaluate our method on the task of climate pattern segmentation. The task is first proposed by 
Mudigonda \etal \cite{mudigonda2017segmenting}, and the goal is to predict extreme weather events, \ie Tropical Cyclones (TC) and Atomospheric Rivers (AT), from simulated global climate data. The training set consists of 43,916 patterns, and 6,274 samples are used for validation. Evaluation results on the validation set are shown in \Tabref{tab:climate_results} and \figref{fig:climate_comparison}.
\begin{table}[h]
    \centering
    \scalebox{0.95}{
    \begin{tabular}{c | c c c | c c}
                Method        &  BG     & TC    & AR    & Mean  & mAP            \\ \hline
                Gauge Net\cite{cohen2019gauge}   &  \bf{97.4}   & \bf{97.9}  & \bf{97.8}  & \bf{97.7}  & \bf{0.759}          \\
                UGSCNN\cite{jiang2019spherical}   &  97.\_  & 94.\_ & 93.\_ & 94.7  & -              \\ \hline
   \textbf{HexRUNet-8}   &  95.71  & 95.57 & 95.19 & 95.49 & 0.518 \\
   \textbf{HexRUNet-32}  &  97.31  & 96.31 & 97.45 & 97.02 & 0.555  \\ \hline
    \end{tabular}}\vspace{0.5em}
    \caption{Climate pattern segmentation results. We include mean class accuracy and mean average precision (mAP) where available. (The background class is denoted BG.)}
    \label{tab:climate_results}
\end{table}
\begin{table*}[t]
    \centering
    \scalebox{0.8}{\begin{tabular}{c | c | c c c c c c c c c c c c c}
                     Method& mIoU    & beam & board & bookcase & ceiling & chair & clutter & column & door & floor & sofa & table & wall & window \\\hline
                     UNet  & 35.9 &  8.5 & 27.2 & 30.7 & 78.6 & 35.3 & 28.8 &  4.9 & 33.8 & 89.1 &  8.2 & 38.5 & 58.8 & 23.9 \\
                     Gauge Net & 39.4 &  --  &  --  & --   & --   & --   & --   & --   & --   &  --  &  --  &  --  &  --  & --   \\
                     UGSCNN & 38.3 &  8.7 & 32.7 & 33.4 & 82.2 & 42.0 & 25.6 & 10.1 & 41.6 & 87.0 &  7.6 & 41.7 & 61.7 & 23.5 \\ \hline
         \textbf{HexRUNet} & \bf{43.3} & \bf{10.9} & \bf{39.7} & \bf{37.2} & \bf{84.8} & \bf{50.5} & \bf{29.2} & \bf{11.5} & \bf{45.3} & \bf{92.9} & \bf{19.1} & \bf{49.1} & \bf{63.8} & \bf{29.4} \\ \hline
    \end{tabular}}\vspace{0.5em}
    \caption{Mean intersection over union (IoU) comparison on 2D3DS dataset. Per-class IoU is shown when available.}
    \label{tab:2d3ds_iou}
\end{table*}
\begin{table*}[t]
    \centering
    \scalebox{0.8}{\begin{tabular}{c | c | c c c c c c c c c c c c c}
                   Method  & mAcc & beam & board & bookcase & ceiling & chair & clutter & column & door & floor & sofa & table & wall & window \\\hline
                     UNet  & 50.8 & 17.8 & 40.4 & 59.1 & 91.8 & 50.9 & \bf{46.0} & 8.7 & 44.0 & 94.8 & 26.2 & 68.6 & 77.2 & 34.8 \\
                     Gauge Net& 55.9 &  --  &  --  & --   & --   & --   & --   & --   & --   &  --  &  --  &  --  &  --  & --   \\
                     UGSCNN & 54.7 & 19.6 & 48.6 & 49.6 & 93.6 & 63.8 & 43.1 & \bf{28.0} & 63.2 & \bf{96.4} & 21.0 & 70.0 & 74.6 & 39.0 \\ \hline
         \textbf{HexRUNet} & \bf{58.6} & \bf{23.2} & \bf{56.5} & \bf{62.1} & \bf{94.6} & \bf{66.7} & 41.5 & 18.3 & \bf{64.5} & 96.2 & \bf{41.1} & \bf{79.7} & \bf{77.2} & \bf{41.1} \\ \hline
    \end{tabular}}\vspace{0.5em}
    \caption{Mean class accuracy (mAcc) comparison on 2D3DS dataset. Per-class accuracy is shown when available.}
    \label{tab:2d3ds_acc}
\end{table*}
Here, we use the same residual U-Net architecture as UGSCNN \cite{jiang2019spherical}. We include two variants using different numbers of parameters: HexRUNet-8 and HexRUNet-32 use 8 and 32 as output channels for the first convolution layer, respectively. As is shown, both versions outperform UGSCNN in terms of mean accuracy. With 32 features, HexRUNet-32's mean accuracy is similar to best performing Gauge Net. However, our method does not match Gauge Net in terms of mean average precision (mAP). We attribute this to the fact that there is no direct orientation information to exploit in this climate data. In contrast, Gauge Net shows its advantage of weight sharing across orientations.

\subsection{Stanford 2D3DS}\label{subsec:2d3ds}
For our first omnidirectional semantic segmentation experiment, we evaluate our method on the 2D3DS dataset \cite{armeni2017joint}, which consists of 1413 equirectangular RGB-D images. The groundtruth 
attributes each pixel to one of 13 classes. Following \cite{jiang2019spherical}, we convert the depth data to be in meter unit and clip to between 0 and 4 meters. RGB data is converted to be in the range of [0, 1] by dividing 255. Finally, all data is mean subtracted and standard deviation normalized. The preprocessed signals are sampled on a level-5 mesh ($r = 5$) using bi-linear interpolation for images and nearest-neighbors for labels. Class-wise weighted cross-entropy loss is used to balance the class examples.

Using our proposed network operators, we employ the residual U-Net architecture of \cite{jiang2019spherical}, which we call HexRUNet (see Sup. Mat. for details). We evaluate our method following the 3-fold splits, and report the mean intersection over union (mIoU) and class accuracy (mAcc) in  \Tabref{tab:2d3ds_iou} and \ref{tab:2d3ds_acc}, respectively.  Our method outperforms orientation-aware method UGSCNN \cite{jiang2019spherical}, rotation-equivariant method Gauge Net \cite{cohen2019gauge} and the U-Net baseline \cite{ronneberger2015u} on equirectangular images that have been sub-sampled to mach level-5 mesh resolution. As for per-class evaluations, our method achieves best performance in most classes. This demonstrates that semantic segmentation indeed benefits from our orientation-aware network.

\begin{figure}[t]
    \centering
    \subfloat{\begin{minipage}{0.8em}\vspace{-5.3em}\rotatebox{90}{RGB}\end{minipage}}\hfill%
    \includegraphics[trim={1cm 1cm 1cm 1cm},clip, width=0.22\linewidth]{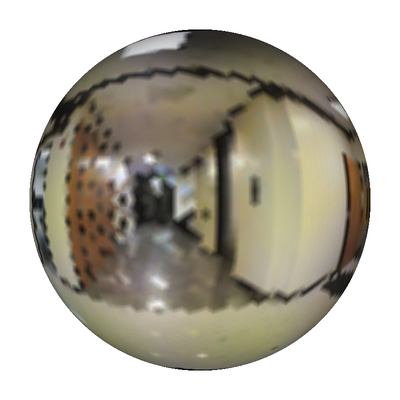}\hfill%
    \includegraphics[trim={1cm 1cm 1cm 1cm},clip, width=0.22\linewidth]{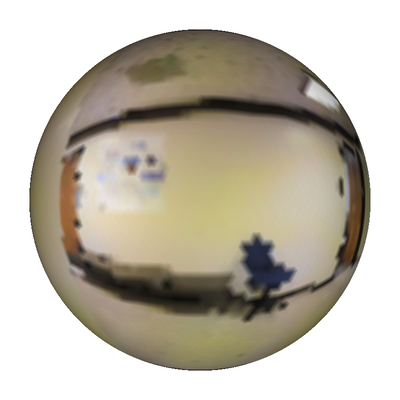}\hfill%
    \includegraphics[trim={1cm 1cm 1cm 1cm},clip, width=0.22\linewidth]{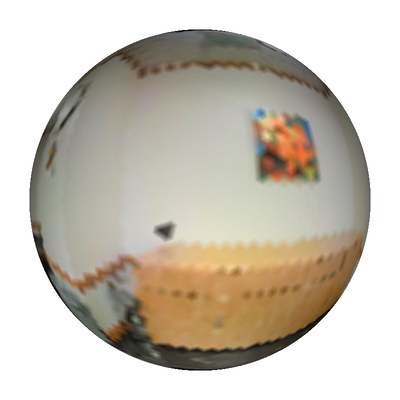}\hfill%
     \includegraphics[trim={1cm 1cm 1cm 1cm},clip, width=0.22\linewidth]{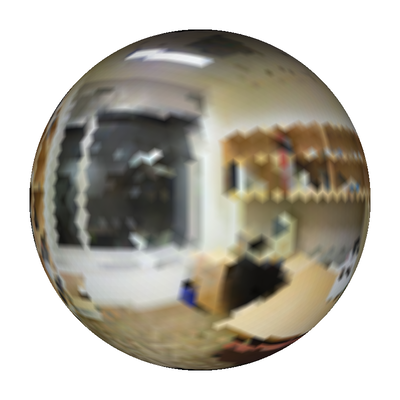}
    \\
     \subfloat{\begin{minipage}{0.8em}\vspace{-5.3em}\rotatebox{90}{GT}\end{minipage}}\hfill%
     \includegraphics[trim={1cm 1cm 1cm 1cm},clip, width=0.22\linewidth]{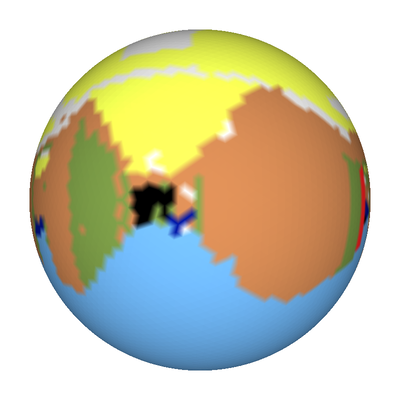}\hfill%
    \includegraphics[trim={1cm 1cm 1cm 1cm},clip, width=0.22\linewidth]{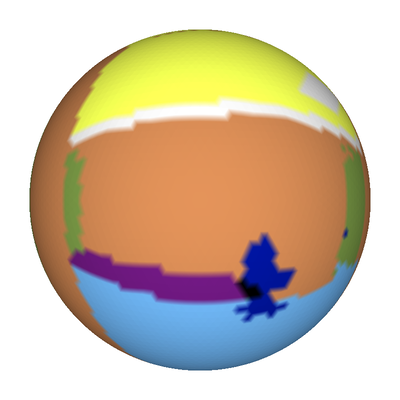}\hfill%
    \includegraphics[trim={1cm 1cm 1cm 1cm},clip, width=0.22\linewidth]{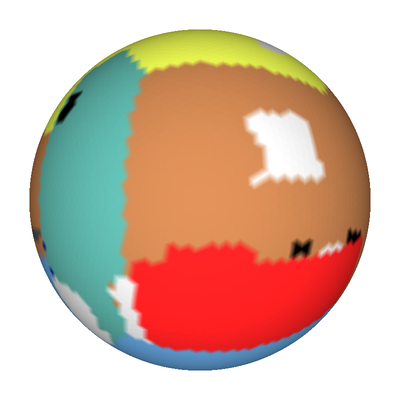}\hfill%
     \includegraphics[trim={1cm 1cm 1cm 1cm},clip, width=0.22\linewidth]{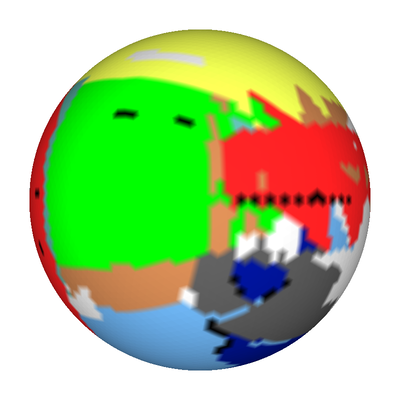}
    \\
    \subfloat{\begin{minipage}{0.8em}\vspace{-5.3em}\rotatebox{90}{UGSCNN}\end{minipage}}\hfill%
     \includegraphics[trim={1cm 1cm 1cm 1cm},clip, width=0.22\linewidth]{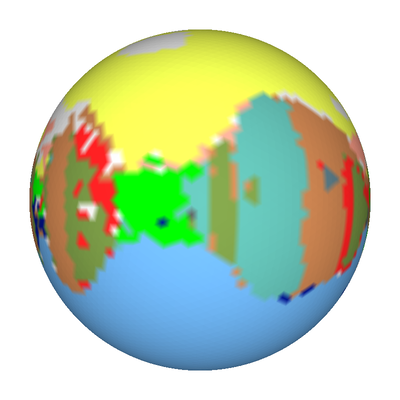}\hfill%
    \includegraphics[trim={1cm 1cm 1cm 1cm},clip, width=0.22\linewidth]{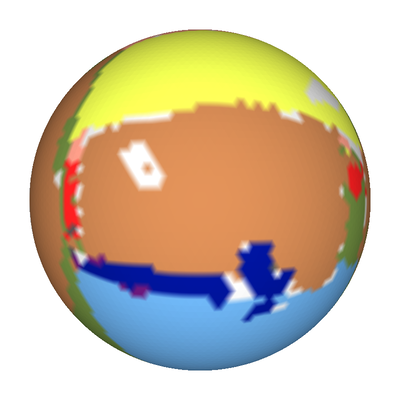}\hfill%
    \includegraphics[trim={1cm 1cm 1cm 1cm},clip, width=0.22\linewidth]{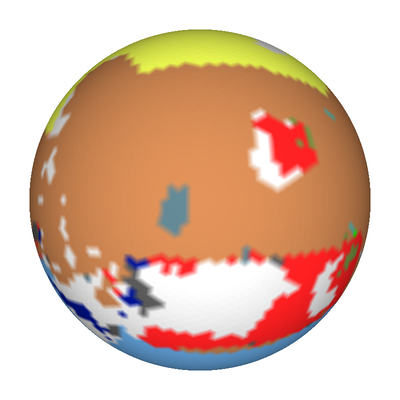}\hfill%
     \includegraphics[trim={1cm 1cm 1cm 1cm},clip, width=0.22\linewidth]{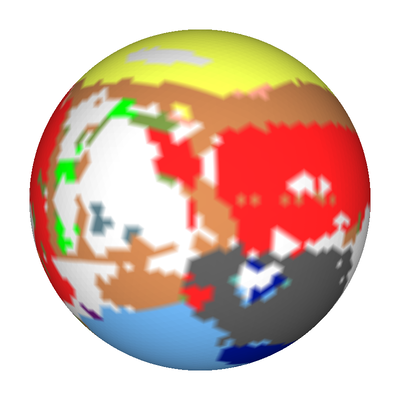}
    \\
    \subfloat{\begin{minipage}{0.8em}\vspace{-5em}\rotatebox{90}{HexRUNet}\end{minipage}}\hfill%
     \includegraphics[trim={1cm 1cm 1cm 1cm},clip, width=0.22\linewidth]{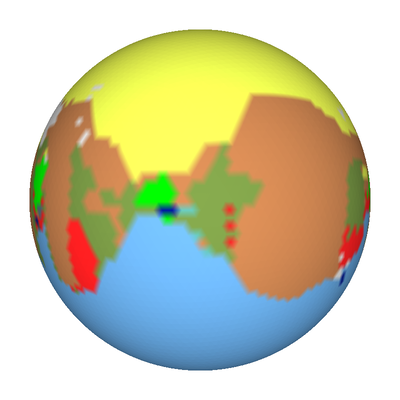}\hfill%
    \includegraphics[trim={1cm 1cm 1cm 1cm},clip, width=0.22\linewidth]{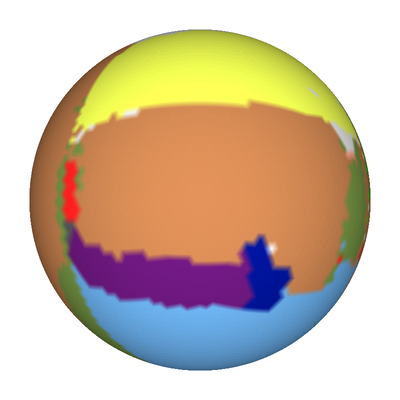}\hfill%
    \includegraphics[trim={1cm 1cm 1cm 1cm},clip, width=0.22\linewidth]{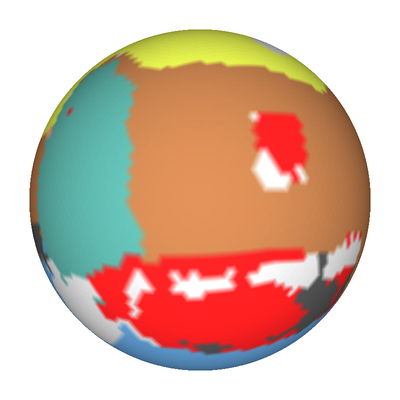}\hfill%
     \includegraphics[trim={1cm 1cm 1cm 1cm},clip, width=0.22\linewidth]{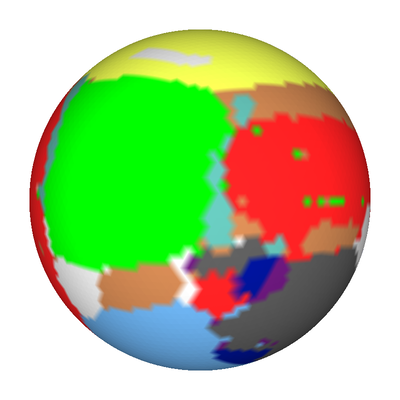}
    \\
    \scalebox{0.75}{\begin{tabular}{l l l l l}
    \fcolorbox{black}{2d3ds-beam}{\rule{0pt}{6pt}\rule{6pt}{0pt}} beam &
    \fcolorbox{black}{2d3ds-board}{\rule{0pt}{6pt}\rule{6pt}{0pt}} board &
    \fcolorbox{black}{2d3ds-bookcase}{\rule{0pt}{6pt}\rule{6pt}{0pt}} bookcase &
    \fcolorbox{black}{2d3ds-ceiling}{\rule{0pt}{6pt}\rule{6pt}{0pt}} ceiling &
    \fcolorbox{black}{2d3ds-chair}{\rule{0pt}{6pt}\rule{6pt}{0pt}} chair \\
    \fcolorbox{black}{2d3ds-clutter}{\rule{0pt}{6pt}\rule{6pt}{0pt}} clutter &
    \fcolorbox{black}{2d3ds-column}{\rule{0pt}{6pt}\rule{6pt}{0pt}} column &
    \fcolorbox{black}{2d3ds-door}{\rule{0pt}{6pt}\rule{6pt}{0pt}} door &
    \fcolorbox{black}{2d3ds-floor}{\rule{0pt}{6pt}\rule{6pt}{0pt}} floor &
    \fcolorbox{black}{2d3ds-sofa}{\rule{0pt}{6pt}\rule{6pt}{0pt}} sofa \\
    \fcolorbox{black}{2d3ds-table}{\rule{0pt}{6pt}\rule{6pt}{0pt}} table &
    \fcolorbox{black}{2d3ds-wall}{\rule{0pt}{6pt}\rule{6pt}{0pt}} wall & 
    \fcolorbox{black}{2d3ds-window}{\rule{0pt}{6pt}\rule{6pt}{0pt}} window &
    \fcolorbox{black}{2d3ds-unknown}{\rule{0pt}{6pt}\rule{6pt}{0pt}} unknown &
    \end{tabular}}\vspace{0.5em}
    \caption{Qualitative segmentation results on 2D3DS dataset.}
    \label{fig:2d3ds_comparison}
    \vspace{-1em}
\end{figure}



\subsection{Omni-SYNTHIA}\label{subsec:synthia}
\begin{table*}[t]
    \centering
    \scalebox{0.8}{\begin{tabular}{c | c | c c c c c c c c c c c c c}
              Method       & mIoU & building & car & cyclist & fence & marking & misc & pedestrian & pole & road & sidewalk & sign & sky & vegetation \\\hline
            UNet    & 38.8 & 80.8 & 59.4 & 0.0 &  0.3 & 54.3 & 12.1 & 4.8 & 16.4 & 74.3 & 58.2 & 0.2 & 90.4 & 49.6 \\
            UGSCNN   & 36.9 & 63.3 & 33.3 & 0.0 &  0.1 & 73.7 &  1.2 & 2.3 & 10.0 & 79.9 & \bf{69.3} & 1.0 & 89.1 & \bf{56.3} \\ \hline
 \textbf{HexUNet-T}  & 36.7 & 71.9 & 53.1 & 0.0 &  1.1 & 69.0 &  4.9 & 0.4 & 11.1 & 72.2 & 52.9 & 0.0 & 92.3 & 48.4 \\
 \textbf{HexUNet-nI} & 42.4 & 77.1 & 64.8 & 0.0 &  2.4 & \bf{74.3} & 10.4 & 2.0 & 23.6 & \bf{84.7} & 68.6 & 1.0 & 93.1 & 48.7 \\
   \textbf{HexUNet}  & \bf{43.6} & \bf{81.0} & \bf{66.9} & 0.0 &  \bf{2.9} & 71.0 & \bf{13.7} & \bf{5.6} & \bf{30.4} & 83.1 & 67.0 & \bf{1.5} & \bf{93.3} & 50.2 \\ \hline
    \end{tabular}}\vspace{0.5em}
    \caption{Mean IoU comparison at $r=6$ on Omni-SYNTHIA dataset.}
    \label{tab:syntia_iou}
\end{table*}
\begin{table*}[t]
    \centering
    \scalebox{0.8}{\begin{tabular}{c | c | c c c c c c c c c c c c c}
               Method      & mAcc & building & car & cyclist & fence & marking & misc & pedestrian & pole & road & sidewalk & sign & sky & vegetation \\\hline
                    UNet    & 45.1 & 91.9 & 63.6 & 0.0 & 4.5 & 57.1 & 17.9 & 5.0 & 19.7 & 88.8 & 73.9 & 0.2 & 94.8 & 69.3 \\
                    UGSCNN   & 50.7 & \bf{93.2} & \bf{81.4} & 0.0 & \bf{5.3} & 83.2 & 33.7 & 2.5 & 14.9 & 90.8 & 82.7 & 1.3 & 96.1 & 74.0 \\ \hline
         \textbf{HexUNet-T}  & 44.8 & 80.0 & 60.9 & 0.0 & 1.6 & 74.7 & 26.9 & 0.4 & 13.0 & 80.0 & 75.2 & 0.0 & \bf{96.2} & 73.4 \\
         \textbf{HexUNet-nI} & 50.6 & 83.9 & 69.6 & 0.0 & 2.5 & 82.9 & \bf{39.1} & 2.0 & 30.7 & \bf{91.8} & 83.6 & 1.1 & 94.8 & \bf{76.5} \\
         \textbf{HexUNet}    & \bf{52.2} & 88.7 & 72.7 & 0.0 & 3.3 & \bf{85.9} & 36.6 & \bf{6.2} & \bf{42.5} & 89.6 & \bf{83.7} & \bf{1.6} & 95.6 & 71.6 \\ \hline
    \end{tabular}}\vspace{0.5em}
    \caption{Per-class accuracy comparison at $r=6$ on Omni-SYNTHIA dataset. }
    \label{tab:syntia_acc}
\end{table*}
To further validate our method on omnidirectional semantic segmentation, we create an omnidirectional version from a subset of the SYNTHIA datset \cite{ros2016synthia}. The SYNTHIA dataset consists of multi-viewpoint photo-realistic frames rendered from a virtual
city and comes with pixel-level semantic annotations for 13 classes. We refer the readers to \cite{ros2016synthia} for details.
We select the ``Summer" sequences of all five places ($2\times$New York-like, $2\times$Highway and $1\times$European-like) to create our own omnidirectional dataset. We split the dataset into a training set of 1818 images (from New York-like and Highway sequences) and use 451 images of the European-like sequence for validation. Only RGB channels are used in our experiments. The icosahedron mesh is populated with data from equirectangular images using interpolation for RGB data and nearest neighbor for labels. Again, we report mIoU and mAcc. Here we use the standard U-Net architecture \cite{ronneberger2015u} to facilitate weight transfer from perspective U-Net in one of our experiments. We call this network ``HexUNet". For an ablation study, we also evaluate our method without north-alignment described in \secref{par:north}, denoted as ``HexUNet-nI".

\begin{figure}[t]
    \centering
    \subfloat{\begin{minipage}{0.8em}\vspace{-5.3em}\rotatebox{90}{RGB}\end{minipage}}\hfill%
    \includegraphics[trim={1cm 1cm 1cm 1cm},clip,width=0.22\linewidth]{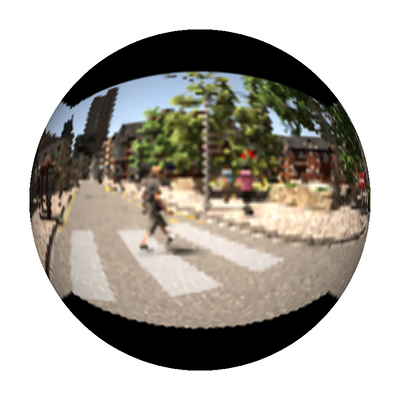}\hfill%
    \includegraphics[trim={1cm 1cm 1cm 1cm},clip,width=0.22\linewidth]{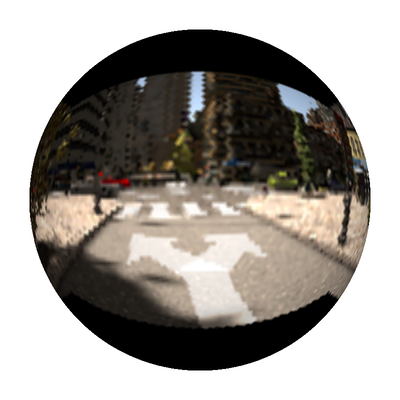}\hfill%
    \includegraphics[trim={1cm 1cm 1cm 1cm},clip,width=0.22\linewidth]{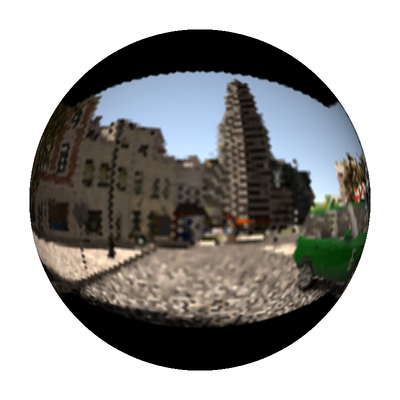}\hfill%
    \includegraphics[trim={1cm 1cm 1cm 1cm},clip,width=0.22\linewidth]{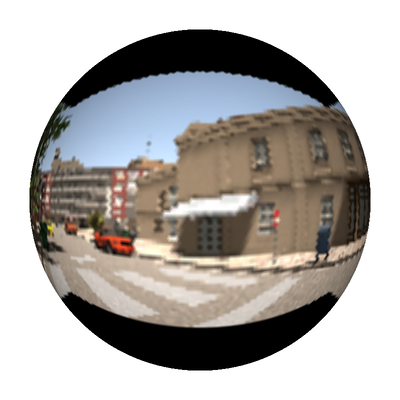} \\
    \subfloat{\begin{minipage}{0.8em}\vspace{-5.3em}\rotatebox{90}{GT}\end{minipage}}\hfill%
     \includegraphics[trim={1cm 1cm 1cm 1cm},clip,width=0.22\linewidth]{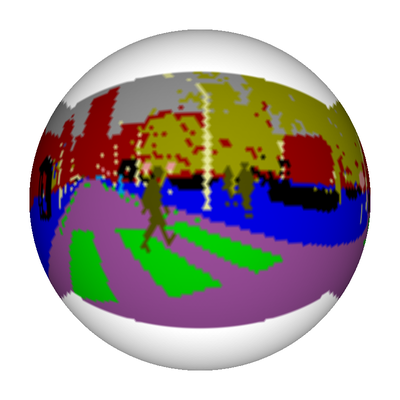}\hfill%
    \includegraphics[trim={1cm 1cm 1cm 1cm},clip,width=0.22\linewidth]{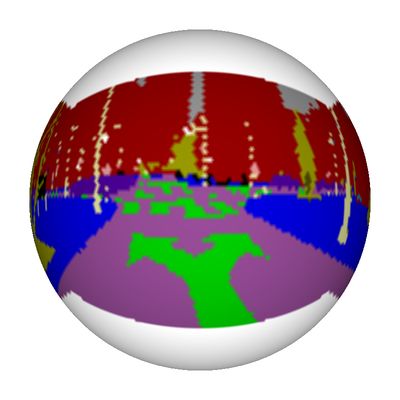}\hfill%
    \includegraphics[trim={1cm 1cm 1cm 1cm},clip,width=0.22\linewidth]{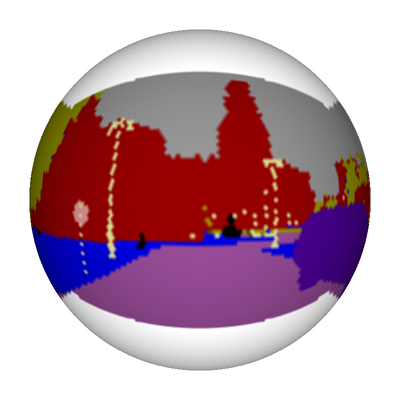}\hfill%
    \includegraphics[trim={1cm 1cm 1cm 1cm},clip,width=0.22\linewidth]{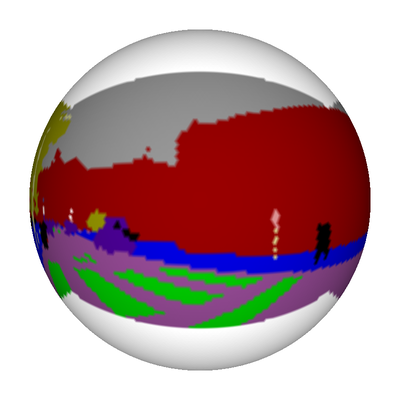} \\
    \subfloat{\begin{minipage}{0.8em}\vspace{-5.3em}\rotatebox{90}{UGSCNN}\end{minipage}}\hfill%
      \includegraphics[trim={1cm 1cm 1cm 1cm},clip,width=0.22\linewidth]{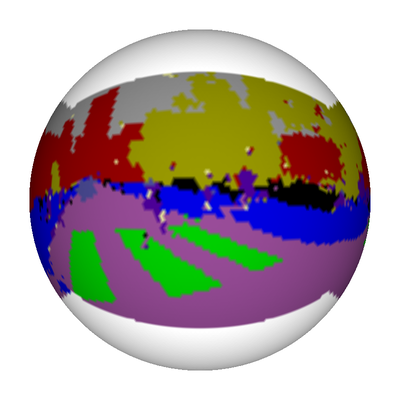}\hfill%
    \includegraphics[trim={1cm 1cm 1cm 1cm},clip,width=0.22\linewidth]{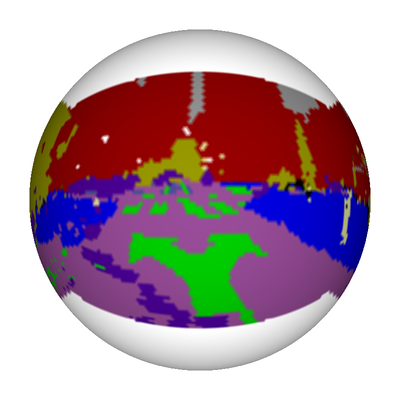}\hfill%
    \includegraphics[trim={1cm 1cm 1cm 1cm},clip,width=0.22\linewidth]{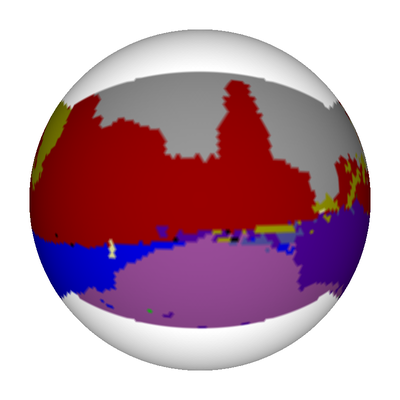}\hfill%
    \includegraphics[trim={1cm 1cm 1cm 1cm},clip,width=0.22\linewidth]{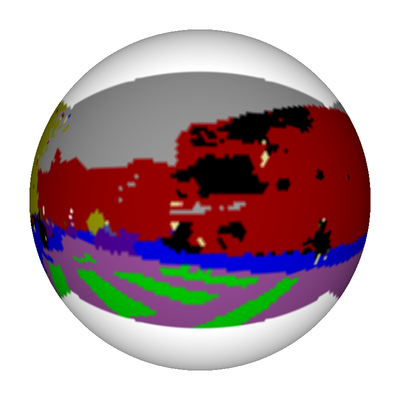} \\
    \subfloat{\begin{minipage}{0.8em}\vspace{-5em}\rotatebox{90}{HexUNet}\end{minipage}}\hfill%
      \includegraphics[trim={1cm 1cm 1cm 1cm},clip,width=0.22\linewidth]{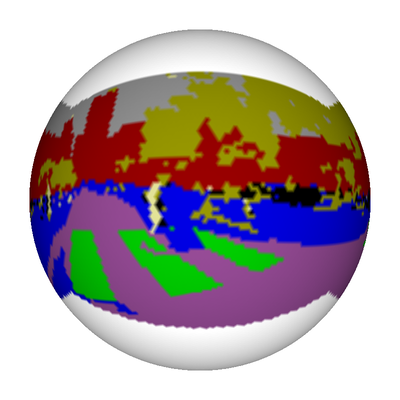}\hfill%
    \includegraphics[trim={1cm 1cm 1cm 1cm},clip,width=0.22\linewidth]{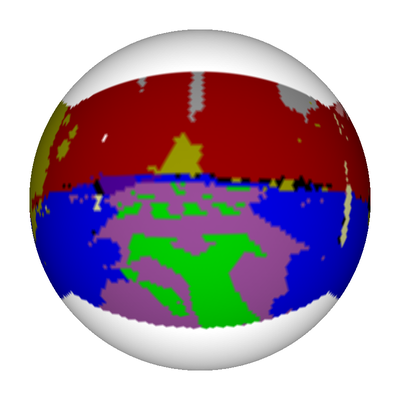}\hfill%
    \includegraphics[trim={1cm 1cm 1cm 1cm},clip,width=0.22\linewidth]{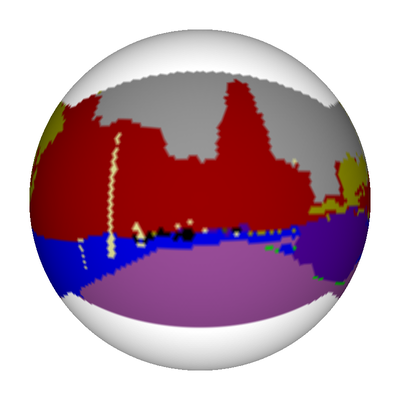}\hfill%
    \includegraphics[trim={1cm 1cm 1cm 1cm},clip,width=0.22\linewidth]{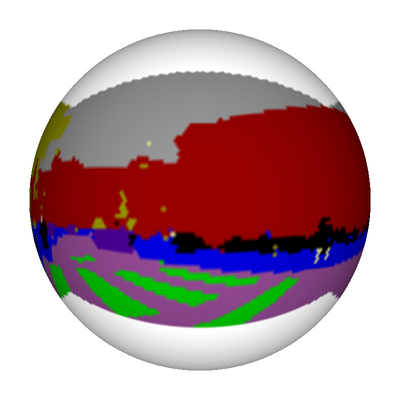} \\
    \scalebox{0.75}{\begin{tabular}{l l l l l}
        \hspace{-0.5em}\fcolorbox{black}{synthia-building}{\rule{0pt}{6pt}\rule{6pt}{0pt}} building &
        \fcolorbox{black}{synthia-car}{\rule{0pt}{6pt}\rule{6pt}{0pt}} car &
        \fcolorbox{black}{synthia-cyclist}{\rule{0pt}{6pt}\rule{6pt}{0pt}} cyclist &
        \fcolorbox{black}{synthia-fence}{\rule{0pt}{6pt}\rule{6pt}{0pt}} fence &
        \fcolorbox{black}{synthia-lanemarking}{\rule{0pt}{6pt}\rule{6pt}{0pt}} marking \\
        \hspace{-0.5em}\fcolorbox{black}{synthia-misc}{\rule{0pt}{6pt}\rule{6pt}{0pt}} misc &
        \fcolorbox{black}{synthia-pedestrian}{\rule{0pt}{6pt}\rule{6pt}{0pt}} pedestrian &
        \fcolorbox{black}{synthia-pole}{\rule{0pt}{6pt}\rule{6pt}{0pt}} pole &
        \fcolorbox{black}{synthia-road}{\rule{0pt}{6pt}\rule{6pt}{0pt}} road &
        \fcolorbox{black}{synthia-sidewalk}{\rule{0pt}{6pt}\rule{6pt}{0pt}} sidewalk \\
        \hspace{-0.5em}\fcolorbox{black}{synthia-sign}{\rule{0pt}{6pt}\rule{6pt}{0pt}} sign &
        \fcolorbox{black}{synthia-sky}{\rule{0pt}{6pt}\rule{6pt}{0pt}} sky & 
        \fcolorbox{black}{synthia-vegetation}{\rule{0pt}{6pt}\rule{6pt}{0pt}} vegetation &
        \fcolorbox{black}{synthia-invalid}{\rule{0pt}{6pt}\rule{6pt}{0pt}} invalid &
        \end{tabular}}\vspace{0.5em}
    \caption{Segmentation results on Omni-SYNTHIA dataset.}
    \vspace{-1em}
    \label{fig:synthia_comparison}
\end{figure}

\paragraph{Comparison with State-of-the-art} We compare our method to UGSCNN \cite{jiang2019spherical} using data sampled at mesh level-6 ($r=6$). We also include planar U-Net \cite{ronneberger2015u} using original perspective images, which have been sub-sampled to match the icosahedron resolution (see Sup. Mat. for details). \Tabref{tab:syntia_iou} and \ref{tab:syntia_acc} report mIoU and mAcc respectively, while \figref{fig:synthia_comparison} shows qualitative results. HexUNet outperforms previous state-of-the-art with significant margin across most classes. The performance on small objects, \eg ``pedestrian" and ``sign", is poor, while all methods fail for ``cyclist". We attribute this to an unbalanced dataset. Note here, class-wise weighted cross-entropy loss is not used. Finally we emphasize, HexUNet performs slightly better than HexUNet-nI, thus verifying the importance of orientation-aware filters in semantic segmentation. 

\begin{table}[t]
    \centering
    \scalebox{0.85}{\begin{tabular}{c | c c | c c | c c}
                        \multirow{2}{*}{Method}    & \multicolumn{2}{c | }{$r = 6$} & \multicolumn{2}{c | }{$r = 7$} & \multicolumn{2}{c}{$r = 8$} \\ 
                            & mIoU & mAcc & mIoU & mAcc & mIoU & mAcc \\\hline
                       UNet & 38.8  & 45.1 & 44.6 & 52.6 & 43.8 & 52.4 \\
                      UGSCNN & 36.9  & 50.7 &   -- &   -- &   -- & --   \\ \hline
         \textbf{HexUNet-T}  & 36.7  & 44.8 & 38.0 & 47.2 & 45.3 & 52.8\\
         \textbf{HexUNet-nI} & 42.4  & 50.6 & 45.1 & 53.4 & 45.4 & 53.2 \\
            \textbf{HexUNet} & \bf{43.6}  & \bf{52.2} & \bf{48.3} & \bf{57.1} & \bf{47.1} & \bf{55.1} \\ \hline
    \end{tabular}}\vspace{0.5em}
    \caption{Evaluation at different resolution on Omni-SYNTHIA dataset. (The current implementation of \cite{jiang2019spherical} could not fit data with resolution higher than $r=6$ to our 11Gb GPU memory, thus no results are available for UGSCNN at $r=\{7, 8\}$.)}
    \label{tab:synthia_detail}
\end{table}
\begin{figure*}[t]
\centering
\subfloat{\begin{minipage}{1em}\vspace{-6.3em}\rotatebox{90}{RGB}\end{minipage}}\hspace{1em}%
\subfloat{\includegraphics[width=0.3\linewidth]{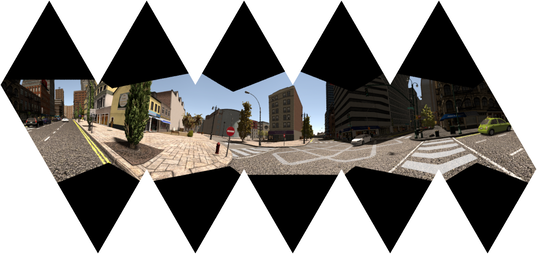}}\hspace{1em}%
\subfloat{\includegraphics[width=0.3\linewidth]{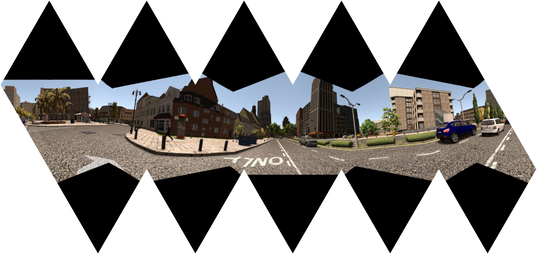}}\hspace{1em}%
\subfloat{\includegraphics[width=0.3\linewidth]{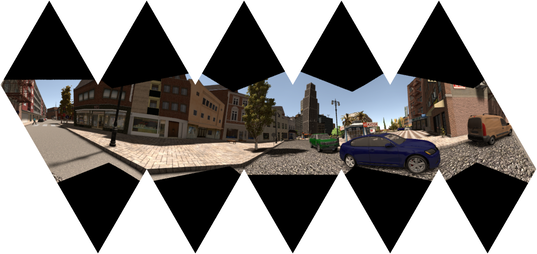}}\vspace{-2.6em}\\
\subfloat{\begin{minipage}{1em}\vspace{-6.3em}\rotatebox{90}{GT}\end{minipage}}\hspace{1em}%
\subfloat{\includegraphics[width=0.3\linewidth]{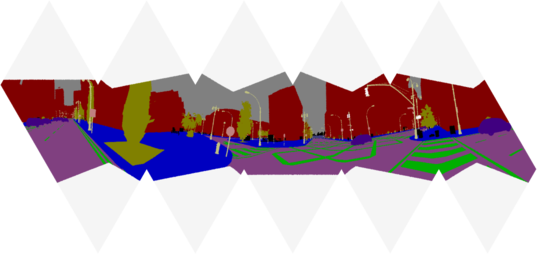}}\hspace{1em}%
\subfloat{\includegraphics[width=0.3\linewidth]{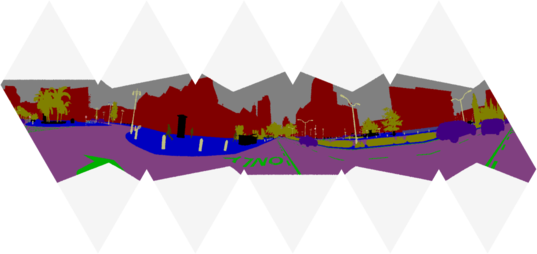}}\hspace{1em}%
\subfloat{\includegraphics[width=0.3\linewidth]{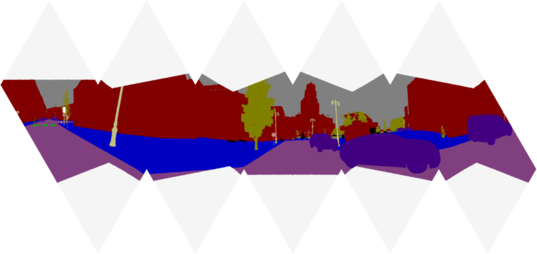}}\vspace{-2.6em}\\
\subfloat{\begin{minipage}{1em}\vspace{-6.3em}\rotatebox{90}{$r = 6$}\end{minipage}}\hspace{1em}%
\subfloat{\includegraphics[width=0.3\linewidth]{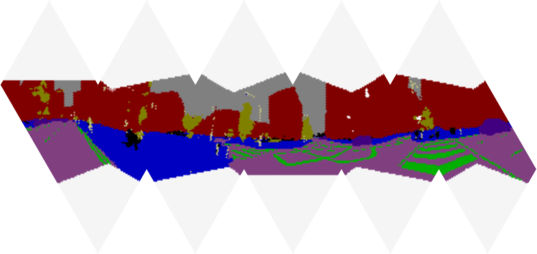}}\hspace{1em}%
\subfloat{\includegraphics[width=0.3\linewidth]{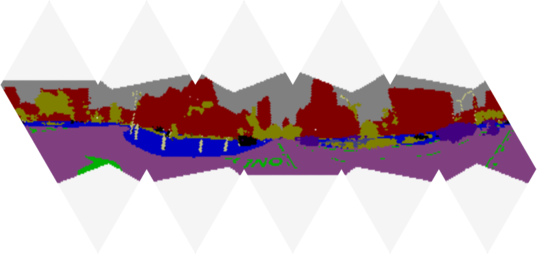}}\hspace{1em}%
\subfloat{\includegraphics[width=0.3\linewidth]{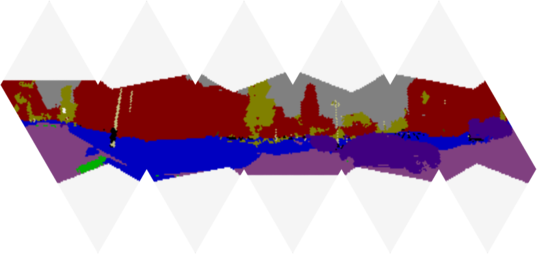}}\vspace{-2.6em}\\
\subfloat{\begin{minipage}{1em}\vspace{-6.3em}\rotatebox{90}{$r = 7$}\end{minipage}}\hspace{1em}%
\subfloat{\includegraphics[width=0.3\linewidth]{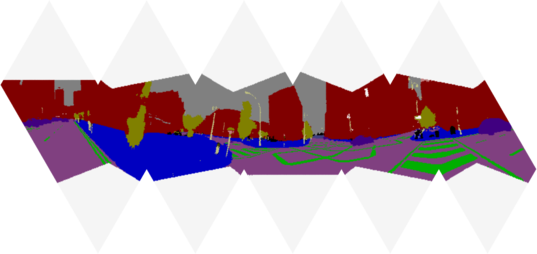}}\hspace{1em}%
\subfloat{\includegraphics[width=0.3\linewidth]{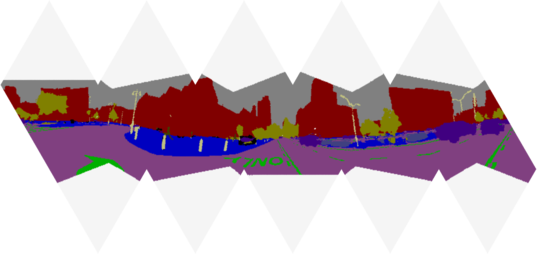}}\hspace{1em}%
\subfloat{\includegraphics[width=0.3\linewidth]{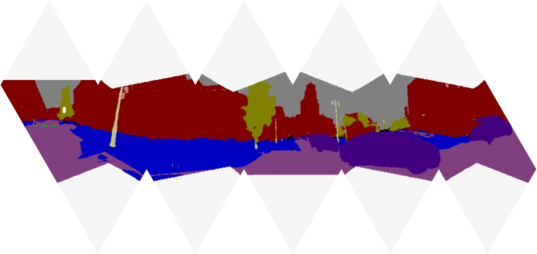}}\vspace{-2.6em}\\
\subfloat{\begin{minipage}{1em}\vspace{-6.3em}\rotatebox{90}{$r = 8$}\end{minipage}}\hspace{1em}%
\subfloat{\includegraphics[width=0.3\linewidth]{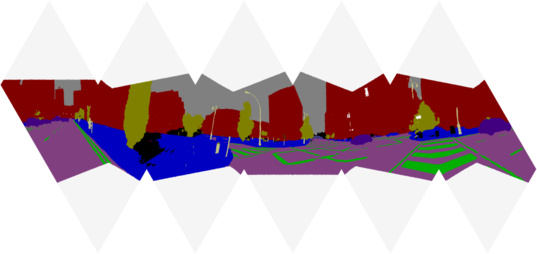}}\hspace{1em}%
\subfloat{\includegraphics[width=0.3\linewidth]{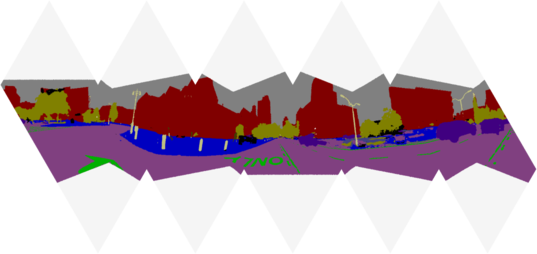}}\hspace{1em}%
\subfloat{\includegraphics[width=0.3\linewidth]{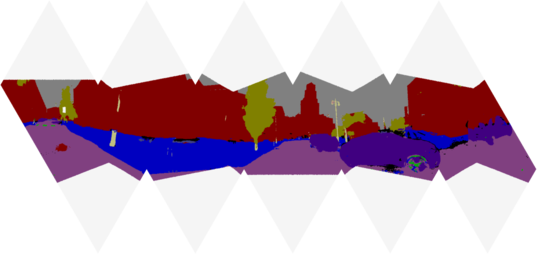}}\\
\scalebox{0.75}{\fcolorbox{black}{synthia-invalid2}{\rule{0pt}{6pt}\rule{6pt}{0pt}} invalid\hspace{0.2em}
        \fcolorbox{black}{synthia-building}{\rule{0pt}{6pt}\rule{6pt}{0pt}} building\hspace{0.2em}
        \fcolorbox{black}{synthia-car}{\rule{0pt}{6pt}\rule{6pt}{0pt}} car\hspace{0.2em}
        \fcolorbox{black}{synthia-cyclist}{\rule{0pt}{6pt}\rule{6pt}{0pt}} cyclist\hspace{0.2em}
        \fcolorbox{black}{synthia-fence}{\rule{0pt}{6pt}\rule{6pt}{0pt}} fence\hspace{0.2em}
        \fcolorbox{black}{synthia-lanemarking}{\rule{0pt}{6pt}\rule{6pt}{0pt}} marking\hspace{0.2em}
        \fcolorbox{black}{synthia-misc}{\rule{0pt}{6pt}\rule{6pt}{0pt}} misc\hspace{0.2em}
        \fcolorbox{black}{synthia-pedestrian}{\rule{0pt}{6pt}\rule{6pt}{0pt}} pedestrian\hspace{0.2em}
        \fcolorbox{black}{synthia-pole}{\rule{0pt}{6pt}\rule{6pt}{0pt}} pole\hspace{0.2em}
        \fcolorbox{black}{synthia-road}{\rule{0pt}{6pt}\rule{6pt}{0pt}} road\hspace{0.2em}
        \fcolorbox{black}{synthia-sidewalk}{\rule{0pt}{6pt}\rule{6pt}{0pt}} sidewalk\hspace{0.2em}
        \fcolorbox{black}{synthia-sign}{\rule{0pt}{6pt}\rule{6pt}{0pt}} sign\hspace{0.2em}
        \fcolorbox{black}{synthia-sky}{\rule{0pt}{6pt}\rule{6pt}{0pt}} sky\hspace{0.2em}
        \fcolorbox{black}{synthia-vegetation}{\rule{0pt}{6pt}\rule{6pt}{0pt}} vegetation}\vspace{0.5em}
\caption{Unfolded visualization of semantic segmentation results at different resolutions on Omni-SYNTHIA dataset. }
\label{fig:resolutions}
\end{figure*}

\paragraph{Evaluation at Different Resolutions} Most previous methods limit their mesh resolution to level $r=5$ which consists of merely 2,562 vertices to represent omnidirectional input. In contrast, an icosahedron mesh at level $r=8$ is required to match the pixel number of $640\times1024$ images, with $655,362\approx 655,360$. Since we believe high resolution input/output is beneficial for the semantic segmentation task, we evaluate our method at different resolutions ($r=\{6,7,8\}$), shown in \tabref{tab:synthia_detail}. Our method achieves best performance at $r=7$, while $r=7$ and $r=8$ perform similar. Since we use a standard U-Net structure consisting of only 4 encoder (and decoder) layers, perception of context is reduced at $r=8$. This is further illustrated by the bottom-rightmost result in \Figref{fig:resolutions}, where a car's wheel is misclassified as road-markings. Resolution $r=6$ and $r=7$ are able to adequately label this. Finally, network training times are shown in \tabref{tab:synthia_runtime}.
\begin{table}[h]
    \centering
    \begin{tabular}{ c| c c c}
                            & $r = 6$ & $r = 7$ & $r = 8$ \\ \hline
                      UGSCNN &   1055s &  7817s  &  --  \\
            \textbf{HexUNet} &   238s  & 406s & 978s \\ \hline
    \end{tabular}\vspace{0.5em}
    \caption{Comparison of average training time per epoch. Evaluations are performed on one Nvidia 1080Ti GPU with 11Gb memory. (Our method is implemented and benchmarked in Tensorflow, while PyTorch implementation of \cite{jiang2019spherical} is used for the comparison.) }
    \label{tab:synthia_runtime}
\end{table}

\paragraph{Evaluation of Perspective Weights Transfer} As shown in \secref{subsec:transfer}, our method utilizes an orientation-aware hexagon convolution kernel which allows direct weight transfer from perspective networks. Initialized with the learned filters ($3\times3$ kernels) from perspective U-Net, we perform weight refinement of only 10 epochs (in contrast to up-to 500 epochs otherwise), and report results as ``HexUNet-T'' in \tabref{tab:syntia_iou}, \ref{tab:syntia_acc} and \ref{tab:synthia_detail}. The proposed filter transfer obtains competitive results, especially at resolution level $r=8$.


%% file: sections/sec5_conclusion.tex
\section{Conclusion}

We introduced a novel method to perform CNN operations on spherical images, represented on an icosahedron mesh. Our method exploits orientation information, as we introduce an efficient interpolation of kernel convolutions, based on north-alignment. The proposed framework is simple to implement, and memory efficient execution is demonstrated for input meshes of level $r = 8$ (equivalent to a $640\times1024$ equirectangular image). In our evaluation on 2D3DS data \cite{armeni2017joint} and our Omni-SYNTHIA version of SYNTHIA \cite{ros2016synthia}, our method becomes the new state of the art for the omnidirectional semantic segmentation task. Furthermore, weight transfer from pretrained standard perspective CNNs was illustrated in our work.

One limitation of the proposed approach is the poor segmentation accuracy for small objects (\eg ``pedestrian" and ``cyclist") which we attribute to unbalanced dataset. 
Future work will incorporate better architectures such as \cite{zhao2017pspnet, sandler2018mobilenetv2} for improved segmentation of small objects. Finally, we plan to exploit our framework for further orientation-aware learning tasks, such as localization and mapping.


%% file: ICCV19.bbl
\begin{thebibliography}{10}\itemsep=-1pt

\bibitem{armeni2017joint}
I.~Armeni, S.~Sax, A.~R. Zamir, and S.~Savarese.
\newblock Joint 2d-3d-semantic data for indoor scene understanding.
\newblock {\em arXiv preprint arXiv:1702.01105}, 2017.

\bibitem{boscaini2016learning}
D.~Boscaini, J.~Masci, E.~Rodol{\`a}, and M.~Bronstein.
\newblock Learning shape correspondence with anisotropic convolutional neural
  networks.
\newblock In {\em NIPS'16}, pages 3189--3197, 2016.

\bibitem{bronstein2017geometric}
M.~M. Bronstein, J.~Bruna, Y.~LeCun, A.~Szlam, and P.~Vandergheynst.
\newblock Geometric deep learning: going beyond euclidean data.
\newblock {\em IEEE Signal Processing Magazine}, 34(4):18--42, 2017.

\bibitem{chen2018deeplab}
L.-C. Chen, G.~Papandreou, I.~Kokkinos, K.~Murphy, and A.~L. Yuille.
\newblock Deeplab: Semantic image segmentation with deep convolutional nets,
  atrous convolution, and fully connected crfs.
\newblock {\em IEEE Trans. Pattern Anal. Mach. Intell.}, 40(4):834--848, 2018.

\bibitem{cheng2018cube}
H.-T. Cheng, C.-H. Chao, J.-D. Dong, H.-K. Wen, and T.-L. Liu.
\newblock Cube padding for weakly-supervised salience prediction in $360^\circ$
  videos.
\newblock In {\em CVPR'19}, 2019.

\bibitem{cohen2018spherical}
T.~S. Cohen, M.~Geiger, J.~K{\"o}hler, and M.~Welling.
\newblock Spherical {CNNs}.
\newblock In {\em ICLR'18}, 2018.

\bibitem{cohen2019gauge}
T.~S. Cohen, M.~Weiler, B.~Kicanaoglu, and M.~Welling.
\newblock Gauge equivariant convolutional networks and the icosahedral {CNN}.
\newblock {\em arXiv preprint arXiv:1902.04615}, 2019.

\bibitem{cohen2016steerable}
T.~S. Cohen and M.~Welling.
\newblock Steerable {CNNs}.
\newblock {\em arXiv preprint arXiv:1612.08498}, 2016.

\bibitem{coors2018spherenet}
B.~Coors, A.~P. Condurache, and A.~Geiger.
\newblock {SphereNet}: Learning spherical representations for detection and
  classification in omnidirectional images.
\newblock In {\em ECCV'18}, pages 518--533, 2018.

\bibitem{dai2017deformable}
J.~Dai, H.~Qi, Y.~Xiong, Y.~Li, G.~Zhang, H.~Hu, and Y.~Wei.
\newblock Deformable convolutional networks.
\newblock In {\em ICCV'17}, pages 764--773, 2017.

\bibitem{esteves2018learning}
C.~Esteves, C.~Allen-Blanchette, A.~Makadia, and K.~Daniilidis.
\newblock Learning so (3) equivariant representations with spherical cnns.
\newblock In {\em ECCV'18}, pages 54 -- 70, 2018.

\bibitem{hoogeboom2018hexaconv}
E.~Hoogeboom, J.~W. Peters, T.~S. Cohen, and M.~Welling.
\newblock Hexaconv.
\newblock {\em arXiv preprint arXiv:1803.02108}, 2018.

\bibitem{jeon2017active}
Y.~Jeon and J.~Kim.
\newblock Active convolution: Learning the shape of convolution for image
  classification.
\newblock In {\em CVPR'17}, pages 4201--4209, 2017.

\bibitem{jiang2019spherical}
C.~M. Jiang, J.~Huang, K.~Kashinath, Prabhat, P.~Marcus, and M.~Nie{\ss}ner.
\newblock Spherical {CNN}s on unstructured grids.
\newblock In {\em ICLR'19}, 2019.

\bibitem{lai2018semantic}
W.-S. Lai, Y.~Huang, N.~Joshi, C.~Buehler, M.-H. Yang, and S.~B. Kang.
\newblock Semantic-driven generation of hyperlapse from 360 degree video.
\newblock {\em IEEE Trans. Visualization and Computer Graphics},
  24(9):2610--2621, 2018.

\bibitem{lee2018spherephd}
Y.~K. Lee, J.~Jeong, J.~S. Yun, C.~W. June, and K.-J. Yoon.
\newblock Spherephd: Applying cnns on a spherical polyhedron representation of
  360 degree images.
\newblock {\em arXiv preprint arXiv:1811.08196}, 2018.

\bibitem{liu2019deep}
M.~Liu, F.~Yao, C.~Choi, S.~Ayan, and K.~Ramani.
\newblock Deep learning 3d shapes using alt-az anisotropic 2-sphere
  convolution.
\newblock In {\em ICLR'19}, 2019.

\bibitem{long2015fully}
J.~Long, E.~Shelhamer, and T.~Darrell.
\newblock Fully convolutional networks for semantic segmentation.
\newblock In {\em CVPR'15}, pages 3431--3440, 2015.

\bibitem{monroy2018salnet360}
R.~Monroy, S.~Lutz, T.~Chalasani, and A.~Smolic.
\newblock Salnet360: Saliency maps for omni-directional images with cnn.
\newblock {\em Signal Processing: Image Communication}, 2018.

\bibitem{monti2017geometric}
F.~Monti, D.~Boscaini, J.~Masci, E.~Rodola, J.~Svoboda, and M.~M. Bronstein.
\newblock Geometric deep learning on graphs and manifolds using mixture model
  cnns.
\newblock In {\em CVPR'17}, pages 5115--5124, 2017.

\bibitem{mudigonda2017segmenting}
M.~Mudigonda, S.~Kim, A.~Mahesh, S.~Kahou, K.~Kashinath, D.~Williams,
  V.~Michalski, T.~O’Brien, and M.~Prabhat.
\newblock Segmenting and tracking extreme climate events using neural networks.
\newblock In {\em Deep Learning for Physical Sciences Workshop, held with
  NIPS'17}, 2017.

\bibitem{ronneberger2015u}
O.~Ronneberger, P.~Fischer, and T.~Brox.
\newblock U-net: Convolutional networks for biomedical image segmentation.
\newblock In {\em MICCAI'15}, pages 234--241, 2015.

\bibitem{ros2016synthia}
G.~Ros, L.~Sellart, J.~Materzynska, D.~Vazquez, and A.~M. Lopez.
\newblock The synthia dataset: A large collection of synthetic images for
  semantic segmentation of urban scenes.
\newblock In {\em CVPR'16}, pages 3234--3243, 2016.

\bibitem{sandler2018mobilenetv2}
M.~Sandler, A.~Howard, M.~Zhu, A.~Zhmoginov, and L.-C. Chen.
\newblock Mobilenetv2: Inverted residuals and linear bottlenecks.
\newblock In {\em CVPR'18}, pages 4510--4520, 2018.

\bibitem{su2017learning}
Y.-C. Su and K.~Grauman.
\newblock Learning spherical convolution for fast features from 360 imagery.
\newblock In {\em NIPS'17}, pages 529--539, 2017.

\bibitem{su2018kernel}
Y.-C. Su and K.~Grauman.
\newblock Kernel transformer networks for compact spherical convolution.
\newblock {\em arXiv preprint arXiv:1812.03115}, 2018.

\bibitem{sun2016design}
Z.~Sun, M.~Ozay, and T.~Okatani.
\newblock Design of kernels in convolutional neural networks for image
  classification.
\newblock In {\em ECCV'16}, pages 51--66, 2016.

\bibitem{weiler20183d}
M.~Weiler, M.~Geiger, M.~Welling, W.~Boomsma, and T.~Cohen.
\newblock 3d steerable cnns: Learning rotationally equivariant features in
  volumetric data.
\newblock In {\em NIPS'18}, pages 10402--10413, 2018.

\bibitem{worrall2018cubenet}
D.~Worrall and G.~Brostow.
\newblock Cubenet: Equivariance to 3d rotation and translation.
\newblock {\em arXiv preprint arXiv:1804.04458}, 2018.

\bibitem{zhao2017pspnet}
H.~Zhao, J.~Shi, X.~Qi, X.~Wang, and J.~Jia.
\newblock Pyramid scene parsing network.
\newblock In {\em CVPR'17}, pages 2881--2890, 2017.

\bibitem{zhou2017oriented}
Y.~Zhou, Q.~Ye, Q.~Qiu, and J.~Jiao.
\newblock Oriented response networks.
\newblock In {\em CVPR'17}, pages 4961--4970, 2017.

\end{thebibliography}
